%% file: article.tex
\documentclass[sn-nature]{sn-jnl}

\usepackage{graphicx}%
\usepackage{multirow}%
\usepackage{amsmath,amssymb,amsfonts}%
\usepackage{amsthm}%
\usepackage{mathrsfs}%
\usepackage{xcolor}%
\usepackage{textcomp}%
\usepackage{manyfoot}%
\usepackage{booktabs}%
\usepackage{hyperref}
\usepackage{algorithm}%
\usepackage{algorithmicx}%
\usepackage{algpseudocode}%
\usepackage{listings}%
\usepackage{array}
\usepackage{booktabs}
\usepackage{makecell}
\usepackage{subcaption}
\usepackage{bm}
\usepackage{pifont}
\usepackage{xltabular, booktabs}

\theoremstyle{thmstyleone}%
\theoremstyle{thmstyletwo}%

\theoremstyle{thmstylethree}%

\begin{document}

\title[ICON decomposition]{ICON decomposition: Multivariate concept auditing reveals shortcuts that neural networks rely on}

\author*[1,2,3]{\fnm{Roshan Prakash} \sur{Rane}}\email{roshan-prakash.rane@uni-tuebingen.de}

\author[4]{\fnm{Marco} \sur{Simnacher}}

\author[4]{\fnm{Manuel} \sur{Pfeuffer}}

\author[1,2,7,8]{\fnm{Marc-Andre} \sur{Schulz}}

\author[1,2]{\fnm{Nys Tjade} \sur{Siegel}}

\author[5]{\fnm{Maximilian} \sur{Dreyer}}

\author[5]{\fnm{Frederik} \sur{Pahde}}

\author[5,6]{\fnm{Wojciech} \sur{Samek}}

\author[4]{\fnm{Sonja} \sur{Greven}}

\author[1,2,7,8]{\fnm{Kerstin} \sur{Ritter}}\email{kerstin.ritter@uni-tuebingen.de}

\affil[1]{\orgdiv{Hertie Institute for AI in Brain Health}, \orgname{University of Tübingen}, \orgaddress{\city{Tübingen}, \country{Germany}}}

\affil[2]{\orgdiv{Department of Psychiatry and Neurosciences}, \orgname{Charité - Universitätsmedizin Berlin}, \orgaddress{\city{Berlin}, \country{Germany}}}

\affil[3]{\orgdiv{Department of Psychology}, \orgname{Humboldt-Universität zu Berlin}, \orgaddress{\city{Berlin}, \country{Germany}}}

\affil[4]{\orgdiv{Chair of Statistics}, \orgname{Humboldt-Universität zu Berlin}, \orgaddress{\city{Berlin}, \country{Germany}}}

\affil[5]{\orgdiv{Department of Artificial Intelligence}, \orgname{Fraunhofer Heinrich Hertz Institute}, \orgaddress{\city{Berlin}, \country{Germany}}}
\affil[6]{\orgdiv{Department of Electrical Engineering and Computer Science}, \orgname{Technische Universität Berlin}, \orgaddress{\city{Berlin}, \country{Germany}}}

\affil[7]{\orgdiv{Tübingen AI Center}, \orgname{University of Tübingen}, \orgaddress{\city{Tübingen}, \country{Germany}}}
\affil[8]{\orgname{German Center for Mental Health (DZPG)}, \orgaddress{\city{Tübingen}, \country{Germany}}}

\abstract{Deep neural networks often exploit spurious associations, a failure known as shortcut learning.
Before deployment, models should therefore be audited for reliance on concepts such as acquisition artifacts and patient demographics.
Current methods, including linear probes and concept activation vectors, measure concept reliance by asking whether each concept is decodable from a layer.
We show that this framing is misleading: decodability reflects not only concept reliance but also correlations among concepts and the outcome in the auditing dataset.
We reframe shortcut auditing as a decomposition of the variance of intermediary representations with Independent Canonical cONcepts (ICON). 
ICON quantifies the share of a layer's variance explained by each concept conditional on all other concepts and the outcome, and also reports the variance left unexplained.
Because ICON explanations are comparable across layers, their trajectories distinguish concepts that drive predictions from those that are merely encoded in the representations.
In simulated data with ground truth, ICON recovers concept importance more accurately than seven baselines.
In skin-cancer and neuroimaging applications, it separates learned shortcuts from merely correlated concepts, which we confirm with independent validation checks.
}


\maketitle

\input{sections/introduction}
\input{sections/results}
\input{sections/discussion}
\input{sections/methods}

\bmhead{Ethics}
This work was carried out under UK Biobank application number 33073.
UK Biobank holds generic research tissue bank approval from the North West Multi-centre Research Ethics Committee (reference 11/NW/0382), granted in 2011 and renewed in 2016 and 2021, under which the analyses reported here were conducted; no separate institutional ethics approval was required.
All UK Biobank participants gave written informed consent at recruitment, covering the use of their data for health-related research and linkage to their health records.
All concepts are used exactly as UK Biobank encodes them, with categories and cohort statistics listed in Extended Data \autoref{tab:ED:neuro_concepts}.
\texttt{Sex} is Data-Field 31, recorded at recruitment from the NHS central registry and in some cases later updated by the participant, so it is not a measure of gender identity; both sexes are represented in every cohort and no analysis is restricted to one sex.
These variables enter the study only as candidate concepts in a variance decomposition: a high ICON score means a model's representation tracks that variable, not that the variable is causally related to the outcome or to the groups it defines.

The ISIC 2019 Challenge images are publicly released and de-identified; ethical approval and participant consent for their collection and release were obtained by the contributing institutions and are described in the publications of the source datasets, for example~\cite{combalia19_isic2019}.
No participant is identifiable from any figure or table in this manuscript.

\bmhead{Data availability}
The ISIC 2019 Challenge training data (BCN\_20000, HAM10000 and MSK) are publicly available from the ISIC Archive under a CC BY-NC licence (\url{https://challenge.isic-archive.com/landing/2019/})~\cite{combalia19_isic2019}, and the annotations of the natural artifacts are distributed with the code of Pahde et al.~\cite{pahde25_thesis}.
UK Biobank data are available to approved researchers through the UK Biobank Access Management System (\url{https://www.ukbiobank.ac.uk}).
Participant-level data, derived participant-level variables and models trained on UK Biobank data are not shared, in line with the UK Biobank access conditions; the aggregate results behind \autoref{fig:neuro} and \autoref{tab:neuro_ood} are included in the code archive~\cite{rane26_iconcode}.
The ToyBrains datasets of Experiment~1 are simulated and can be regenerated with ToyBrains v1.1.0~\cite{roshanrane_toybrains_2024} and the configuration files in the code archive.

\bmhead{Code availability} 
The ICON implementation and the analysis code for all three experiments are archived at Zenodo (v1.0.0, Apache 2.0 licence)~\cite{rane26_iconcode} and maintained at \url{https://github.com/RoshanRane/ICON_decomposition}.
Experiment~1 uses ToyBrains v1.1.0 (\url{https://github.com/RoshanRane/toybrains/releases/tag/v1.1.0}), and Experiment~2 uses the code of Pahde et al.~\cite{pahde25_thesis} at tag \texttt{icon-v1.0.0} of \url{https://github.com/RoshanRane/medical-ai-safety}.
The ICON implementation, the ToyBrains and UK Biobank experiments and all explanation analyses were run in Python 3.12 with PyTorch 2.11.0 and PyTorch Lightning 2.6.1; the ISIC~2019 dataset generation and VGG-16 training (Experiment 2) were run in a separate Python 3.10 environment (PyTorch 2.11.0) with the pinned code of Pahde et al.; the learning rate and weight decay of the two UK Biobank models were tuned with Optuna 4.8.0~\cite{akiba2019optuna}, while the ToyBrains and ISIC models used fixed hyperparameters. ICON extends the partial least squares implementation in scikit-learn 1.8.0~\cite{scikit-learn} and uses NumPy 2.4.3, SciPy 1.17.1 and pandas 3.0.2. The conditional independence tests were run in R 4.3.3 with the comets package 0.2-2~\cite{kook24_comets}, whose nuisance regressions are ranger 0.17.0 random forests. T1-weighted images were used as distributed by the UK Biobank imaging pipeline~\cite{alfaro21_ukbbconfound}; locally, they were only cropped and resampled to $96\times114\times96$ voxels. Conda specifications of the three environments used (Python 3.12, Python 3.10 for the ISIC training, and R 4.3.3) are included in the released code, together with exports of the environments in which the reported results were produced.

\bmhead{Use of large language models}
We used Claude from Anthropic for restructuring and copy editing the manuscript text.
All suggestions were checked and, where adopted, revised by the authors, who take full responsibility for the content.

\bmhead{Funding}
Funded by the Deutsche Forschungsgemeinschaft (DFG, German Research Foundation) – 459422098 (Fusing Deep Learning and Statistics towards Understanding Structured Biomedical Data; KI-FOR 5363 DeSBi) to R.P.R., M.S., M.P., M.D., F.P., W.S., S.G., K.R.; 402170461 (Losing and Regaining Control over Drug Intake; TRR 265) to R.P.R., K.R.; 414984028 (SFB 1404 FONDA) to K.R.; 565356445 (Validating Explainable AI in Clinical Neuroimaging) to M.-A.S., N.T.S.; 586414057 (Diagnostic Fairness: Disentangling Data Quality, Confounding, and Algorithmic Bias in Biomedical Regression) to M.-A.S. 
The authors R.P.R., M.-A.S., N.T.S. and K.R. gratefully acknowledge funding by Gemeinnützige Hertie Stiftung and Cluster of Excellence ``Machine Learning -- New Perspectives for Science''. This research has been conducted using the UK Biobank Resource under application number 33073.

\bmhead{Author contributions}
R.P.R.: conceptualization, methodology, investigation, software, visualization, writing -- original draft, writing -- review and editing.
M.S.: conceptualization, methodology, investigation, writing -- review and editing.
M.P.: conceptualization, methodology, writing -- review and editing.
M.-A.S.: conceptualization, investigation.
N.T.S.: conceptualization.
M.D.: conceptualization, methodology, investigation.
F.P.: methodology, writing -- review and editing.
W.S.: supervision, writing -- review and editing.
S.G.: supervision, methodology, writing -- review and editing.
K.R.: supervision, methodology, investigation, writing -- review and editing.
All authors read and approved the final manuscript.

\bmhead{Acknowledgements}
We thank the UK Biobank participants and the ISIC Archive contributors, without whom this work would not be possible.

\bmhead{Competing interests}
The authors declare no competing interests.

\bibliography{bibliography}

\input{sections/supplementary}

\input{sections/extended_data}

\end{document}

%% file: sections/introduction.tex

\begin{figure}[!htbp]
    \centering
    \includegraphics[trim=0.75cm 0 0 0, width=1.04\textwidth]{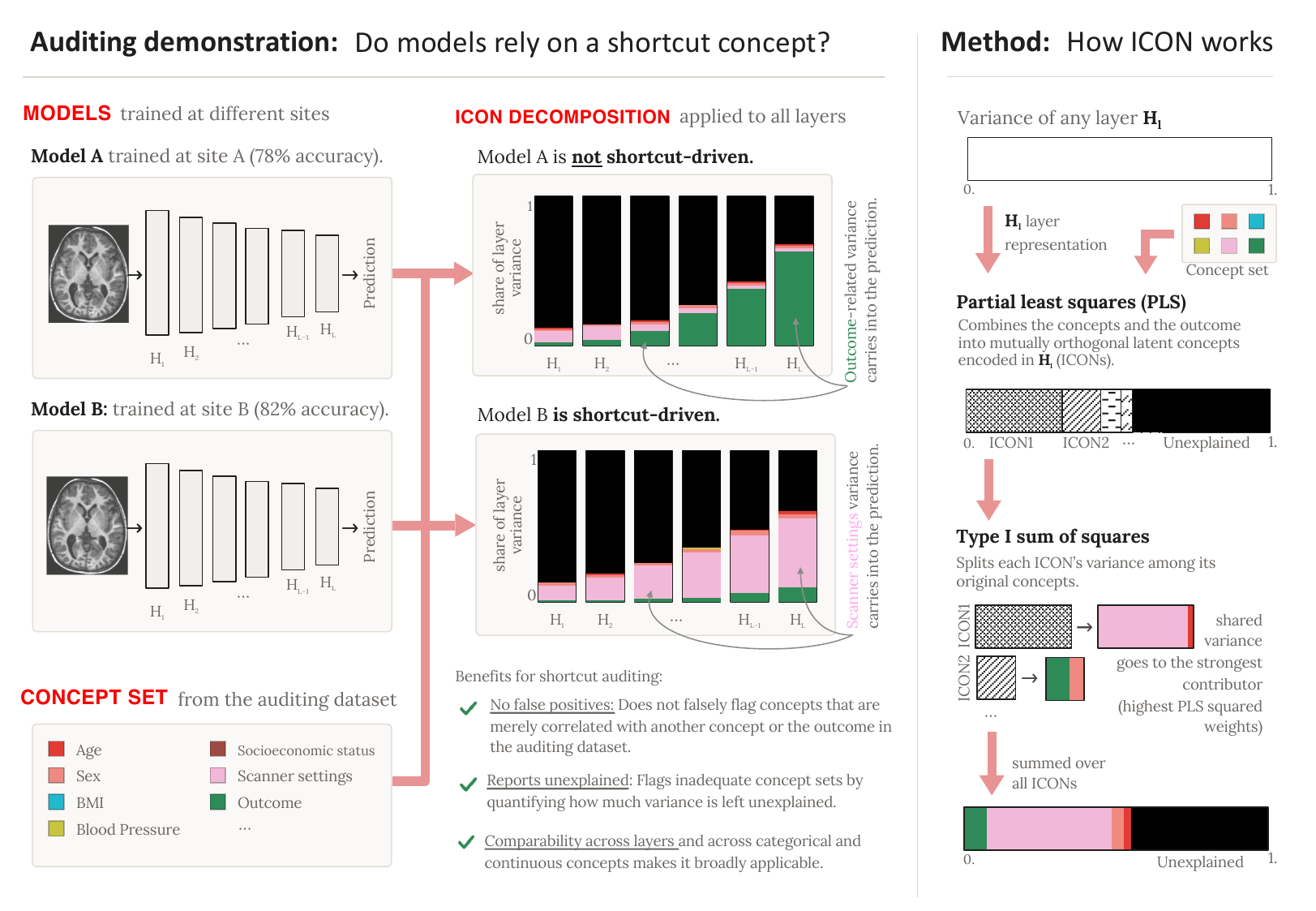}
    \caption{\textbf{Overview of ICON decomposition.}
    \textbf{Left,} a schematic demonstrating how ICON can be used to audit models for shortcuts. Given a trained model, an auditing dataset and a set of concepts, ICON splits the variance of every layer representation into the share explained by each concept, the share explained by the outcome, and the share that none of the provided concepts explains (black). ICON shows model~A is not shortcut-driven  while model~B's predictions can be explained mainly by the \texttt{scanner setting} shortcut, even though the model~B has the higher accuracy.
    \textbf{Right,} the internal mechanics of the ICON method consist of two steps. First, PLS combines the provided concepts and the outcome into mutually orthogonal, latent components encoded in the layer representation, called the Independent Canonical cONcepts (ICONs). A Type~I sum-of-squares allocation then assigns each ICON's variance back to the original concepts, assigning any shared variance to the strongest contributor using the PLS weights. The allocations are summed across all ICONs to obtain the per-concept importance at a layer.}
    \label{fig:abstract}
\end{figure}

Deep neural networks (DNNs) are increasingly employed across biomedicine, from high-stakes clinical decision support to fundamental biomedical research \cite{ramsundar19_DLinbiomedical}.
However, a key barrier to their adoption is their tendency to exploit spurious associations and biases in the training data, a behaviour known as `shortcut learning' \cite{geirhos20_shortcutlearninginDL}.
For example, it has been shown that skin-cancer detection models can rely on skin markings or band-aids rather than on lesions~\cite{winkler19_ISICconfounder, pahde23_reveal}, while brain-disease models can rely on participants' sex or recruitment site rather than on disease-related biomarkers~\cite{thibeau22_adniscanner, rane2022structural}.
Therefore, before deployment, models should be screened for reliance on shortcut concepts such as data-acquisition artifacts and sociodemographic biases~\cite{poeta23_ConceptXAIsurvey}.
This is currently performed using concept-based explainability (C-XAI) methods such as linear probes \cite{alain16_probing} and Concept Activation Vectors (CAVs) \cite{kim18_TCAV, pahde24_signalCAV, graziani2018_RCAV}.
Linear probes and CAVs are popular because they can be applied post-hoc to any trained model and any concept, provided that sufficient concept-annotated samples are available in an auditing dataset. They do not require model retraining or access to the original training data.
However, several studies show that they often produce incorrect and misleading explanations~\cite{elazar21_probingamnesic, ravichander20_probingcorrelation, nicolson25_CAVcritique, dreyer24_hopetosafety, raman24_CAVcorrelation, li24_CAVfaithfulness, ramaswamy23_CAVtestdata, brown21_CAVadversarial} (reviewed in Supplementary Note 1). 
We argue that these failures arise from their framing: they pose C-XAI as a \emph{univariate decoding} problem, asking ``is this concept predictable from a DNN layer's representation?'', and analyse each concept in isolation. 
Three major issues follow from this.

(i) \emph{Univariate decoding conflates concept importance with correlations in the auditing dataset, producing false positives}.
For example, in neuroimaging cohorts, clinical, demographic and socioeconomic variables are strongly interdependent and associated with both the imaging measurements and the outcome~\cite{hyatt20_covarreview, alfaro21_ukbbconfound}.
A univariate probe or CAV can assign high importance to a concept such as \texttt{age} or \texttt{income} simply because it correlates with the outcome, even when the model does not use it as a shortcut~\cite{ravichander20_probingcorrelation, adebayo22_validateTCAV}.
The same occurs between concepts: if a model relies on \texttt{sex}, correlated concepts such as \texttt{income} or \texttt{BMI} are also falsely assigned high importance.

(ii) \emph{Univariate decoding cannot quantify whether the provided concepts are adequate to explain a representation}.
An explanation can only be as good as the concept set it is provided.
Even if \{\texttt{sex}, \texttt{age}, \texttt{income}\} all score highly for a brain-disease model, it may still rely on a genuine biomarker beyond them. 
Scoring concepts individually cannot quantify what the provided concept set accounts for as a whole, and so cannot warn a user when that set is inadequate.

Finally, (iii) \emph{Existing C-XAI scores are not comparable across layers of the same model, nor across categorical and continuous concepts, making them unwieldy as shortcut auditing tools}.
Several studies report incoherent interpretations across layers, with a concept appearing important at one layer but not the next~\cite{nicolson25_CAVcritique, dreyer24_hopetosafety, belinkov22_probingsurvey}. Additionally, different metrics are used for categorical and continuous concepts (accuracy versus $R^2$ for probes~\cite{rane23_deeprepviz}, and testing with CAVs (TCAV)~\cite{kim18_TCAV} versus regression CAVs (RCAV)~\cite{graziani2018_RCAV} for CAVs), with no principled way to reconcile them~\cite{nicolson25_CAVcritique}.

Several fixes have been proposed, such as using covariance-based estimators rather than decoders~\cite{pahde24_signalCAV}, forcing concept orthogonality~\cite{erogullari25_disentangleCAVs}, and adding follow-up diagnostics~\cite{nicolson25_CAVcritique} (Supplementary Note 1).
Each of these, however, fixes one symptom without addressing the incorrect underlying univariate framing that causes these issues.

Therefore, we propose to fundamentally reframe C-XAI as a \emph{multivariate variance-decomposition} problem using our Independent Canonical cONcept (ICON) decomposition method (\autoref{fig:abstract}). 
Rather than asking whether a concept is decodable, ICON asks: ``how much of the variance in a DNN layer's representation is explained by each concept, given the other concepts and the outcome?''
ICON builds on established principles of variance-partitioning~\cite{gromping15_statsvarimp, clouvel24_statsvarimp, dinga20_ypreddecomposition}, combining partial least squares (PLS) to handle high-dimensional representations~\cite{wold1966_plsoriginal} with a Type~I (sequential) sum-of-squares allocation \cite{kutner05_statsbook} to reduce false positives from correlated concepts  (\autoref{fig:abstract}, right).

This reframing addresses all three limitations.
For (i), ICON accounts for correlations among concepts, and with the outcome, explicitly, allocating shared variance to the variable that contributes most strongly.
A concept that is merely correlated with the outcome or another important concept therefore receives little or no importance.
For (ii), ICON reports the share of a representation's variance that \emph{none} of the provided concepts explains.
ICON resolves (iii) by expressing importance as a proportion of representation variance. 
Scores are therefore comparable across layers, since the explained and unexplained shares always sum to one regardless of dimensionality, and one common metric covers categorical and continuous concepts alike.
A large unexplained share is itself informative, as it indicates that the model might have learned a shortcut concept that is missing from the provided concept set, or that the provided concepts are encoded too nonlinearly for a linear method to capture.
ICON explanations therefore carry an explicit assessment of whether the provided concept set is adequate, which no univariate score offers.

We validate ICON in three experiments:
\begin{enumerate}
    \item On the ToyBrains synthetic image benchmark~\cite{roshanrane_toybrains_2024}, where ground-truth concept importance is known, ICON recovers it more accurately than classical variance-decomposition methods, probes, or CAVs. ICON's advantage widens as the auditing dataset shrinks or the concept set grows.
    \item On the ISIC~2019 skin-cancer benchmark~\cite{combalia19_isic2019}, we insert synthetic artifacts at controlled rates and train models to use them as shortcuts. ICON assigns near-zero importance to artifacts the model was never exposed to, and tracks the strength of the shortcut as we increase it. In contrast, competing C-XAI methods produce false positives driven by correlations in the auditing dataset.
    \item Finally, on two open neuroimaging research questions, ICON identifies \texttt{sex} as a shortcut in the first one and surfaces no substantial shortcut concept beyond the outcome in the second. Existing methods produce diffuse and misleading explanations, whereas ICON produces explanations that are sparse and \emph{actionable}, that is, specific enough to motivate follow-up experiments. 
\end{enumerate}

%% file: sections/results.tex

\section{Results} \label{results}
\subsection{In simulated data, ICON is the most accurate method because it rejects spuriously correlated concepts}

\begin{figure*}[!htbp]
    \centering
    \includegraphics[width=0.9\textwidth]{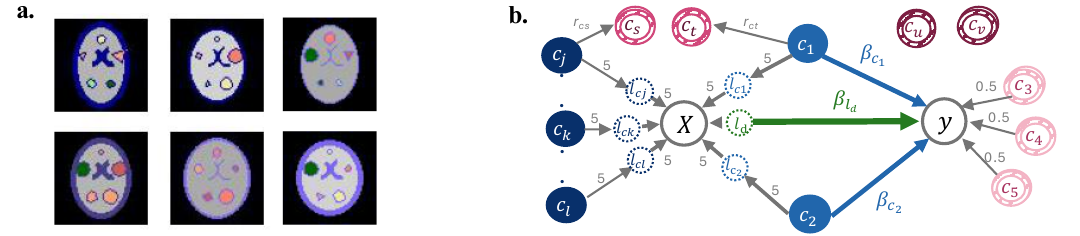}
    \\[0.2em]
    \includegraphics[width=0.9\textwidth]{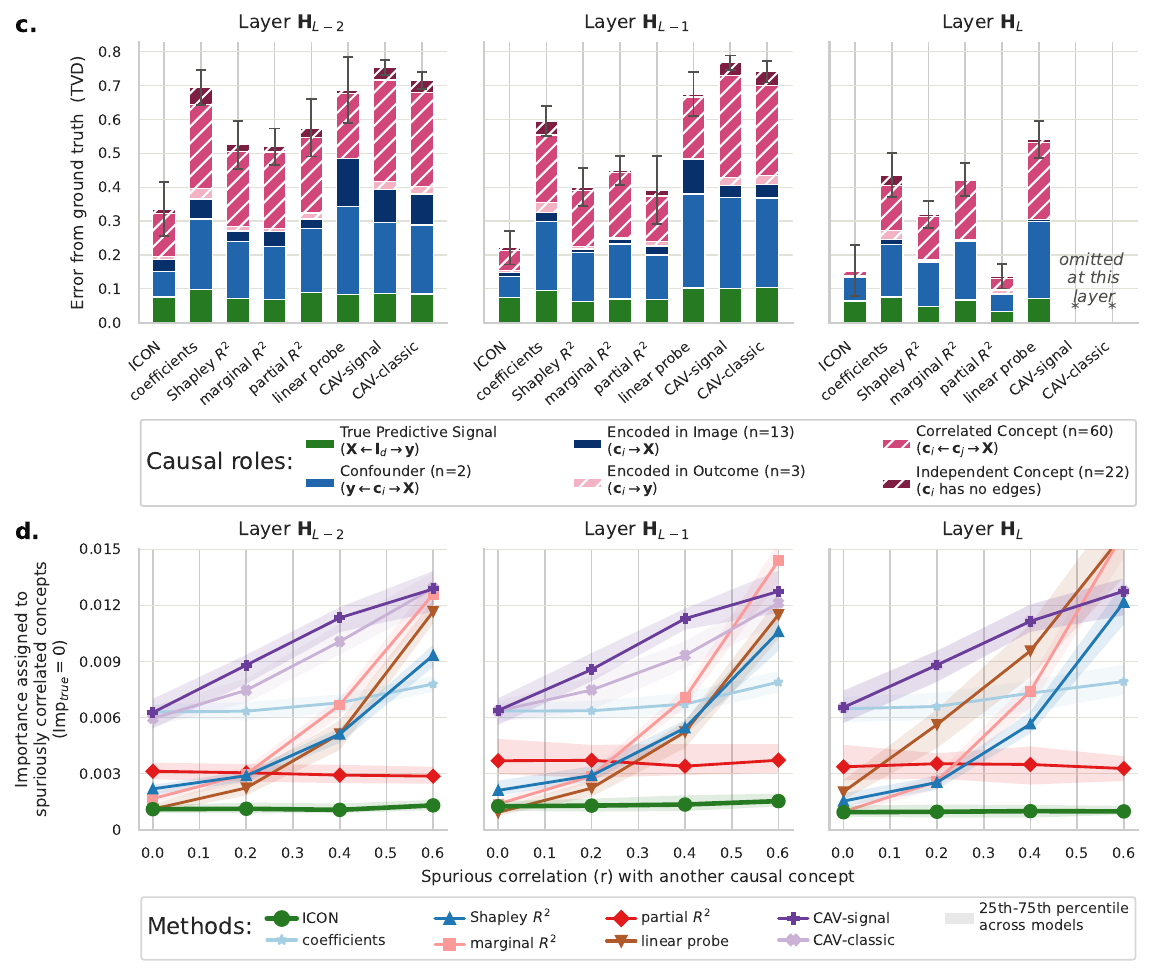}
\caption{\textbf{Simulation dataset setup and results.}
    \textbf{a,}~Example $64\times64$ ToyBrains images, rendered from sixteen image attributes $\mathbf{l}_k$ that set the position, colour and intensity of different shapes.
    \textbf{b,}~Data-generating graph: attributes $\mathbf{l}_k$ (dotted circles) render the image $\mathbf{X}$. The true signal $\mathbf{l}_d$ (green) emulates an unannotated disease biomarker, which every method can approximate only through the outcome $\mathbf{y}$. Confounders $\mathbf{c}_1$, $\mathbf{c}_2$ (blue) emulate shortcuts such as sex or site.
    We tune the edges $\beta_{\mathbf{l}_d}$, $\beta_{\mathbf{c}_1}$, $\beta_{\mathbf{c}_2}$ into $\mathbf{y}$ to generate six scenarios, ranging from purely confounder-driven to mainly signal-driven (Extended Data \autoref{tab:ED:toybrains_methods-settings}). 
    \textbf{c-d,}~Concept importance at the last three CNN layers: $\mathbf{H}_{L-2}$, $\mathbf{H}_{L-1}$ and the logit $\mathbf{H}_{L}$ (architecture in Extended Data \autoref{fig:ED:toybrains-pqsweep}).
    We report medians over $90$ trained models (6 scenarios $\times$ 5 architectures $\times$ 3 seeds), using the same $n = 500$ random test samples and $q = 30$ random concepts for all methods. Error bars and bands denote the 25th to 75th percentiles. 
    We compare ICON to seven baselines (Methods~\ref{sec:methods:baselines}): four classical variable-importance estimators~\cite{gromping15_statsvarimp} (marginal, partial and Shapley $R^2$; standardised coefficients) and three C-XAI methods (linear probes, CAV-classic, CAV-signal).
    \textbf{c,}~Total variation distance (TVD) from ground truth, by concept role. Pink: importance given to truly irrelevant concepts (false positives). Asterisks: CAVs are undefined at the one-dimensional $\mathbf{H}_L$.
    \textbf{d,}~Importance assigned by different methods to spuriously correlated concepts at different correlation strengths $r$ ($r = 0$ line marks independent concepts).
    }\label{fig:res-toybrains}
\end{figure*}

Obtaining the ground-truth importance of individual concepts in a model's representations is difficult in real-world data.
We therefore use the ToyBrains simulator~\cite{roshanrane_toybrains_2024}, where we can specify the causal graph that generates both the images and the outcome (\autoref{fig:res-toybrains}a). We obtain the ground-truth importance of each concept by intervening on this graph and measuring the share of each layer's variance that the intervention accounts for (Methods~\ref{sec:methods:exp1}).
We compare this ground truth against ICON and seven baselines using total variation distance (TVD; $0$ indicates exact recovery, $1$ complete disagreement, \autoref{eq:tvd}). 
The goal of this simulation experiment is to test why ICON's PLS and Type~I sum-of-squares machinery is necessary beyond standard statistical variable-importance estimators~\cite{gromping15_statsvarimp}.

Our causal graph links the image and the outcome through one `true' signal and two confounders (\autoref{fig:res-toybrains}b).  The confounders are part of a pool of $100$ concepts that occupy one of five causal roles (legend of \autoref{fig:res-toybrains}c).
Only $15$ of these causally influence the model representations: $13$ are image-encoded concepts and two are confounders.
The remaining $85$ concepts, represented as pink hatches in \autoref{fig:res-toybrains}b,c, have a ground-truth importance of exactly zero: three influence only the outcome, while the rest can be spuriously correlated with one of the causal concepts at $r \in \{0.0, 0.2, 0.4, 0.6\}$ ($r=0$ implies they are not correlated with any other variable in the graph).

\textbf{ICON has the lowest error in the high-dimensional internal layers, which are the target of C-XAI methods (\autoref{fig:res-toybrains}c).}
ICON's mean TVD error is $0.34$ at $\mathbf{H}_{L-2}$ and $0.22$ at $\mathbf{H}_{L-1}$, compared with $0.52$ and $0.39$ for the best baseline at each layer (marginal $R^2$ and partial $R^2$, respectively).
By contrast, partial $R^2$ slightly outperforms ICON at the scalar decision logit $\mathbf{H}_{L}$ ($0.14$ versus $0.15$), where the one-dimensional representation makes ordinary multiple regression well conditioned.

\textbf{ICON's advantage comes from suppressing the false positives from spuriously correlated concepts and accurately estimating the confounders (\autoref{fig:res-toybrains}c).}
For every baseline, most of the error at $\mathbf{H}_{L-2}$ and $\mathbf{H}_{L-1}$ comes from the confounders (medium-blue) and the spuriously correlated concepts (medium-pink hatch), which together account for $65$--$79\%$ of the TVD. 
The misallocation for the spuriously correlated concepts grows with the spurious correlation $r$ for the univariate baselines (\autoref{fig:res-toybrains}d): from $r = 0$ to $r = 0.6$, the importance of zero-importance concepts rises by a factor of $8$--$12\times$ for the linear probe and $8$--$16\times$ for marginal $R^2$, but only $1.0$--$1.2\times$ for ICON.
Like ICON, partial $R^2$ and the regression coefficients also do not show a dose response, but they already assign $3$--$7\times$ ICON's importance to zero-importance concepts at $r = 0$.

\textbf{ICON's advantage widens at the internal layers as the concept set grows or the auditing dataset shrinks (Extended Data \autoref{fig:ED:toybrains-pqsweep}).} 
As the number of provided concepts grows from $q = 5$ to $100$, ICON's error stays $\approx0.20$ at $\mathbf{H}_{L-1}$, whereas every baseline's error rises to $0.40$--$0.85$ at $q = 100$ (Extended Data \autoref{fig:ED:toybrains-pqsweep}a).
ICON's error also levels off earliest, at about $n = 250$ samples (Extended Data \autoref{fig:ED:toybrains-pqsweep}b). This matters in practice because concept annotations are often available for only a few hundred samples, e.g., in biomedical cohorts. In contrast, the closest contender, partial $R^2$, requires about $n = 5{,}000$ samples to converge to ICON's error level at $\mathbf{H}_{L-1}$. 


\subsection{In skin-cancer classifiers, ICON separates between learned shortcuts and concepts that are merely correlated in the auditing data}

\begin{figure}[!htbp]
    \centering
    \includegraphics[width=0.85\textwidth]{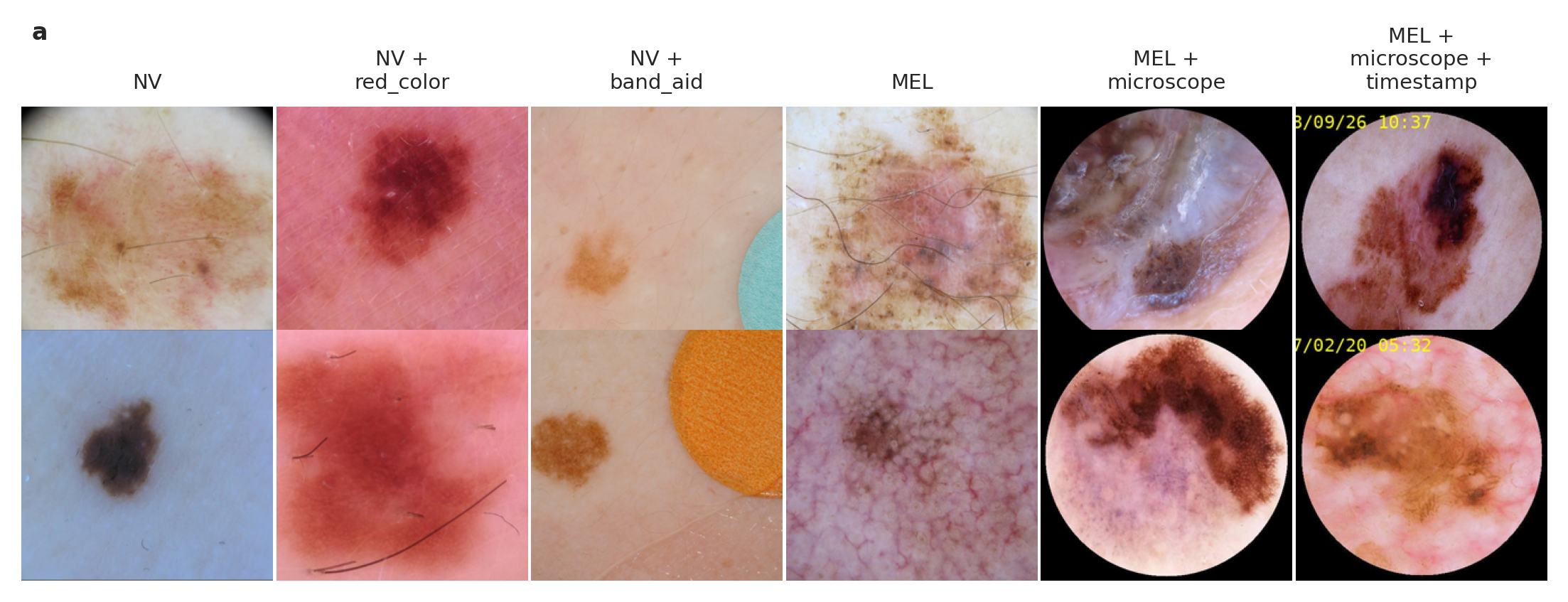}
    \includegraphics[width=\textwidth]{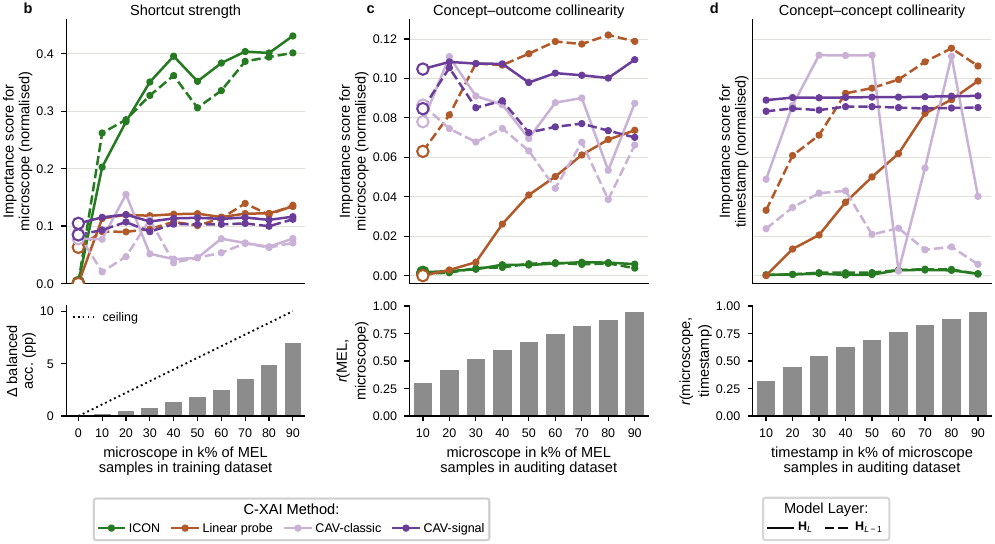}
    \caption{\textbf{ICON's importance for a shortcut concept increases with the model's reliance on it and stays near zero for concepts that the model did not learn.}
    \textbf{a,}~Example images of the nevus (\texttt{NV}) and melanoma (\texttt{MEL}) classes with natural (\texttt{red\_color}, \texttt{band\_aid}) and inserted (\texttt{microscope}, \texttt{timestamp}) artifacts.
    \textbf{b--d,}~Normalised importance of the inserted artifact at the last two layers of the VGG-16 model (top) and the reference quantity that changes in each test (bottom).
    Circles in \textbf{b} and \textbf{c} mark the clean model $\mathcal{M}_{\text{micro-00}}$, which never saw any \texttt{microscope} artifact during training.
    \textbf{b,}~We train ten models $\mathcal{M}_{\text{micro-00}}$ to $\mathcal{M}_{\text{micro-90}}$ and $\Delta\text{acc}$ (bottom) shows the portion of each model's balanced accuracy that is due to its reliance on the \texttt{microscope} shortcut (Methods~\ref{sec:methods:exp2}). All models are audited using the test split of $\mathcal{D}_{\text{micro-10}}$ (\texttt{microscope} present in $10\%$ of the \texttt{MEL} class).
    \textbf{c,}~The clean model $\mathcal{M}_{\text{micro-00}}$, audited with different datasets with \texttt{microscope} in $k\%$ of the \texttt{MEL} class  (bottom, the resulting \texttt{microscope}--\texttt{MEL} Pearson correlation $r$).
    \textbf{d,}~The shortcut-affected model $\mathcal{M}_{\text{micro-40}}$ audited for \texttt{timestamp}, an artifact absent from all training sets, on datasets with \texttt{timestamp} in $k\%$ of the \texttt{microscope} images (bottom, the resulting \texttt{timestamp}--\texttt{microscope} Pearson correlation $r$).
    In both \textbf{c} and \textbf{d}, the expected importance at both layers is zero. 
    Non-normalised scores are shown in Extended Data \autoref{fig:isic_results_abs}b--d.}
    \label{fig:isic_results}
\end{figure}

\begin{figure}[!htbp]
    \centering
    \small
    \setlength{\tabcolsep}{2pt}
    \begin{tabular}{cc}
        \includegraphics[width=\textwidth]{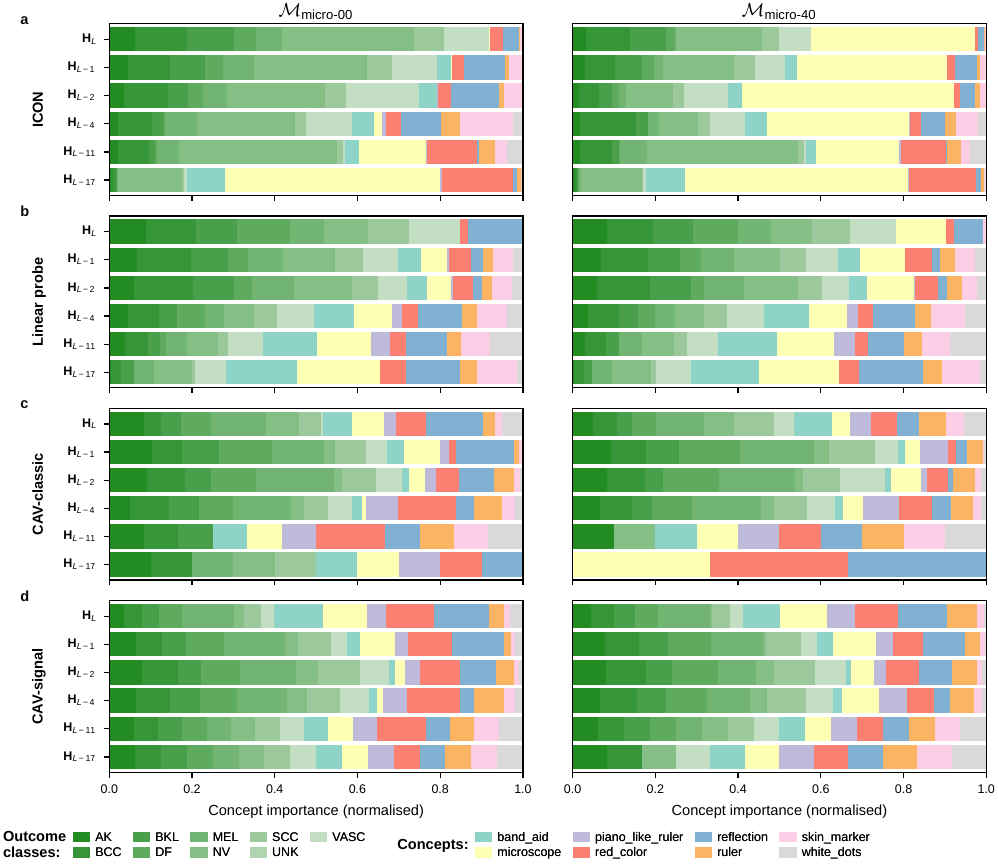}
    \end{tabular}
    \caption{\textbf{ICON's layer-wise decomposition separates a clean model from a shortcut-affected one, whereas the baselines show subtle and inconsistent differences.}
    Rows are the C-XAI methods (\textbf{a},~ICON; \textbf{b},~linear probe; \textbf{c},~CAV-classic; \textbf{d},~CAV-signal), and each stacked bar is that method's normalised importance (\autoref{eq:relimp}) for the nine outcome classes (shades of green) and the nine artifact concepts (other colours).
    Layers go from an early convolutional layer $\mathbf{H}_{L-17}$ (bottom) to the prediction logits $\mathbf{H}_L$ (top).
    Within a row, the \textbf{left panel} is the clean model $\mathcal{M}_{\text{micro-00}}$ and the \textbf{right panel} the shortcut-affected model $\mathcal{M}_{\text{micro-40}}$, both evaluated on the same auditing dataset in which the \texttt{microscope} artifact (yellow) is present in $10\%$ of the \texttt{MEL} images ($n = 2{,}459$ images).
    The non-normalised ICON scores, with the unexplained share retained, are shown in Extended Data \autoref{fig:isic_results_abs}a.}
    \label{fig:isic_qualitative}
\end{figure}

The ISIC 2019 skin cancer dataset~\cite{combalia19_isic2019} is a popular benchmark for evaluating C-XAI methods~\cite{poeta23_ConceptXAIsurvey, dreyer24_hopetosafety, pahde24_signalCAV, nicolson25_CAVcritique} because its images contain artifacts such as surgical skin markings, band-aids, rulers, and camera reflections that are known to act as shortcuts in skin-cancer detection models~\cite{winkler19_ISICconfounder, pahde23_reveal}.

We perform a controlled experiment by inserting two synthetic artifacts into the images at controlled rates $k$: a \texttt{microscope} lens and a \texttt{timestamp} (rightmost examples in \autoref{fig:isic_results}a; Methods~\ref{sec:methods:exp2}).
E.g., $\mathcal{D}_{\text{micro-}k}$ denotes an adulterated variant of the dataset where we inserted the \texttt{microscope} artifact into $k\%$ of the samples of the melanoma class (\texttt{MEL}). 
A DNN model (VGG-16) trained on this adulterated dataset, denoted $\mathcal{M}_{\text{micro-}k}$, learns to rely on the \texttt{microscope}  artifact as a shortcut for \texttt{MEL}, to a degree that grows with $k$ (\autoref{fig:isic_results}b, bottom).

We compare ICON with linear probes, CAV-classic~\cite{kim18_TCAV} and CAV-signal~\cite{pahde24_signalCAV} by normalising their concept importance scores over the concept set  (\autoref{eq:relimp}), which contains the nine artifact concepts and nine outcome classes (Extended Data \autoref{fig:ED:isic_data_dist}).

\textbf{ICON's layer-wise decomposition clearly distinguishes a shortcut-affected model from a clean model (\autoref{fig:isic_results}b, \autoref{fig:isic_qualitative}).}
Across the ten models audited $\mathcal{M}_{\text{micro-00}}$ to $\mathcal{M}_{\text{micro-90}}$, ICON's relative importance for \texttt{microscope} in the concept set rises from $\approx0\%$ to $43\%$ in the prediction logit (\autoref{fig:isic_results}b).
ICON's sensitivity to detect shortcut-affected models is more apparent when we compare its layer-wise decomposition of a shortcut-affected model, $\mathcal{M}_{\text{micro-40}}$, and a clean model, $\mathcal{M}_{\text{micro-00}}$.
ICON reveals the learned \texttt{microscope} shortcut in $\mathcal{M}_{\text{micro-40}}$ as a clear distinction between the two models, whereas the other three C-XAI methods show only subtle and inconsistent differences.
ICON shows that \texttt{microscope} is encoded in the early convolutional layers of both models since the artifact is present in the auditing dataset's input images. But from $\mathbf{H}_{L-4}$ onwards, \texttt{microscope} is filtered out by $\mathcal{M}_{\text{micro-00}}$, while in $\mathcal{M}_{\text{micro-40}}$, \texttt{microscope} occupies a large share up to the prediction logit.

\textbf{Probes and CAVs flag artifacts the model never learned, whereas ICON correctly assigns them near-zero importance (\autoref{fig:isic_results}c,d).}
We audit the clean model $\mathcal{M}_{\text{micro-00}}$ on auditing datasets in which the \texttt{MEL}--\texttt{microscope} correlation rises from $0.30$ to $0.94$ (\autoref{fig:isic_results}c, bottom). 
Since $\mathcal{M}_{\text{micro-00}}$ never saw \texttt{microscope} during training, the model cannot have learned to rely on it.
For all nine auditing datasets, ICON assigns $<1\%$ importance to \texttt{microscope} in the last two layers.
The linear probe's score, however, rises with the \texttt{MEL}--\texttt{microscope} correlation, reaching up to $12\%$ of the concept set, and the two CAV scores stay high and drift non-monotonically. This problem also holds for their non-normalised scores (Extended Data \autoref{fig:isic_results_abs}c).
Next, we repeat this test for a \texttt{timestamp} artifact using the shortcut-affected $\mathcal{M}_{\text{micro-40}}$ and a set of auditing datasets containing a \texttt{timestamp} artifact, such that the \texttt{microscope}--\texttt{timestamp} correlation increases from $0.32$ to $0.95$ (\autoref{fig:isic_results}d, bottom). Because \texttt{timestamp} appears in no training set, $\mathcal{M}_{\text{micro-40}}$ cannot have learned to rely on it.
ICON's importance for \texttt{timestamp} again stays below $1\%$, whereas the linear probe tracks the collinearity in the auditing dataset. CAV-signal stays at $\approx9\%$, and CAV-classic is the least stable, swinging between $0.5\%$ and $11\%$ across adjacent auditing datasets, consistent with previous reports of CAV instability~\cite{ramaswamy23_CAVtestdata, brown21_CAVadversarial}.

We next apply ICON to two open neuroimaging questions in which the true concept importances are unknown.


\subsection{In biomedical research, ICON produces accurate insights about what the model learned}

\begin{figure}[!htbp]
    \centering
    \includegraphics[width=0.96\linewidth]{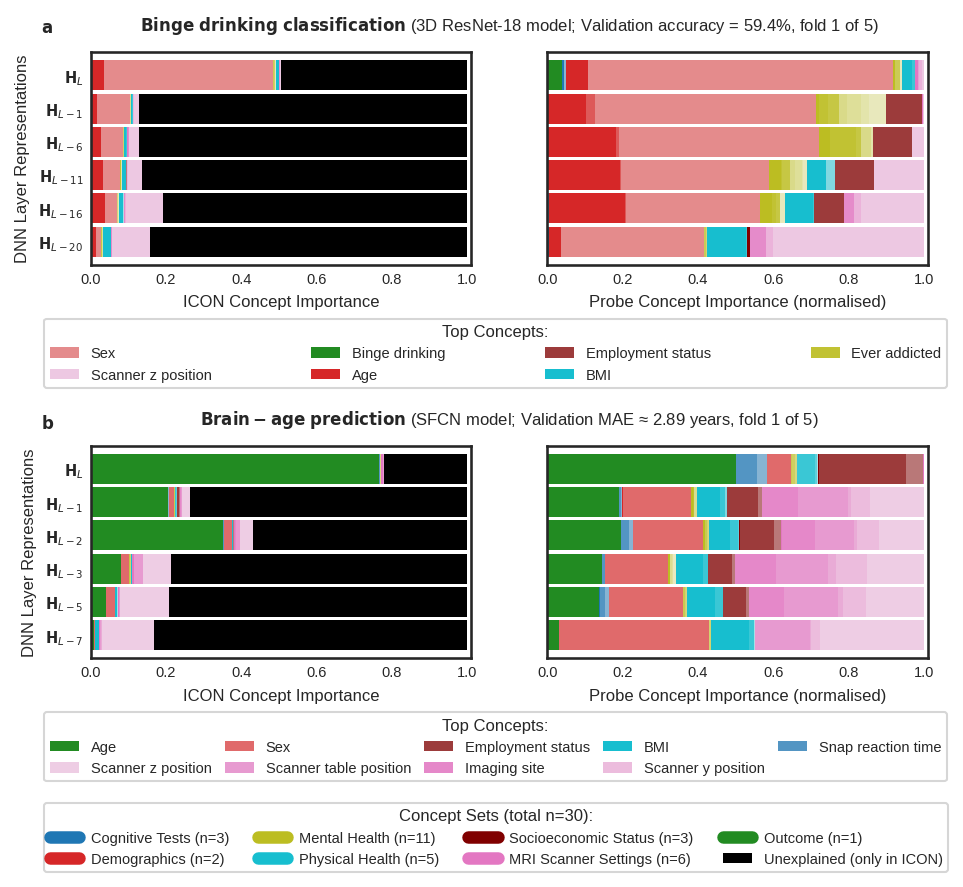}
    \caption{\textbf{On two brain-MRI models, ICON gives sparse and clear explanations, whereas the linear probe's explanations are diffuse and ambiguous.}
    Stacked bars show the importance of the $30$ concepts and the outcome in each interpreted layer, from the earliest (bottom) to the prediction logit $\mathbf{H}_L$ (top).
    ICON \textbf{(left)} and linear probe \textbf{(right)} are computed on the auditing dataset ($n = 1{,}246$ held-out participants in \textbf{a}, $n = 5{,}000$ in \textbf{b}).
    ICON bars are shares of representation variance that sum to one at each layer, with the black segment denoting the share not assigned to any provided concept or the outcome. 
    Probe bars are per-concept decoding scores ($R^2$ or Cohen's $\kappa$) normalised for display, and unlike ICON they carry no unexplained term.
    Colours mark different concepts and are shaded by their category (categories shown in the bottom legend). For each task, concepts with more than $5\%$ importance at any layer are listed below.
    \textbf{a,}~Binge-drinking classification (3D ResNet-18, fold 1 of 5): ICON attributes the final layer mainly to \texttt{sex}, whereas the probe also surfaces the outcome and several other health-related concepts (\texttt{BMI}, \texttt{ever\_addicted}).
    \textbf{b,}~Brain-age prediction (SFCN, fold 1 of 5): ICON attributes the final layer mainly to \texttt{age}, with scanner-related concepts only in early layers, whereas the probe surfaces many other concepts, including \texttt{employment\_status}, \texttt{BMI}, \texttt{sex}, \texttt{snap\_reaction\_time}, and scanner settings.}
    \label{fig:neuro}
\end{figure}

DNNs are increasingly used to find disease biomarkers from high-dimensional biomedical data~\cite{ritter2021_DLinPsy, ramsundar19_DLinbiomedical}, and C-XAI methods can indicate whether these models learn a biologically relevant signal or rely on a sociodemographic bias or an artifact.
We investigate two such neuroimaging biomarkers:
\begin{enumerate}
    \item \textbf{Binge drinking:} An alcohol-misuse phenotype called binge drinking is predictable from structural brain MRI in a European \textit{adolescent} cohort~\cite{rane2022structural, rane23_eating}. We test whether a similar MRI signature can be found in \textit{middle-aged and older} UK Biobank participants~\cite{sudlow15_ukbb}.
    \item \textbf{Brain-age gap:} The brain-age gap is the difference between a person's age and the age a DNN predicts from their brain MRI.
    Researchers hypothesise that this gap captures accelerated brain ageing caused by diseases~\cite{cole19_brainagereview, franke19_brainagereview}.
    We test whether brain-age predictions rely on spurious associations with scanner settings or sociodemographic factors that could compromise the gap as a biomarker.
\end{enumerate}

In biomedicine, candidate concepts are numerous and correlated (Extended Data \autoref{fig:ED:neuro_corr}), precisely the regime where univariate C-XAI methods produce false positives. 
We compare ICON with linear probes (CAVs are excluded; Methods~\ref{sec:methods:exp3}) and use independent tests to resolve disagreements between the two.
Both models are audited with a set of $30$ concepts (Extended Data \autoref{tab:ED:neuro_concepts}), plus the outcome, which serves as a proxy for any biological features learned beyond the provided concepts (Methods~\ref{sec:methods:math}). 

\textbf{On the binge-drinking model, ICON identifies \texttt{sex} as a shortcut, whereas the probe's explanations are ambiguous (\autoref{fig:neuro}a).}
A 3D ResNet-18~\cite{hara18_3dresnets} distinguishes binge drinkers from controls with an accuracy of $60.0 \pm 0.5\%$ (mean $\pm$ s.d. across five folds; chance $50\%$), suggesting a modest predictive signal.
ICON shows that the share of variance explained by \texttt{sex} grows steadily from the earliest layer to the prediction logit $\mathbf{H}_L$, where it is larger than that of any other concept. 
The outcome ($r(\text{outcome},\texttt{sex}) \approx 0.23$) explains essentially no additional variance in $\mathbf{H}_L$ once \texttt{sex} is accounted for.
The probe also ranks \texttt{sex} highly but additionally assigns importance to the outcome and to several health-related concepts (\autoref{fig:neuro}a, right).
This diffuse attribution leaves it unclear whether the model relies only on \texttt{sex} or also captures a genuine binge-drinking signature.

\textbf{Retraining on sex-balanced data supports ICON's reading, as most of the above-chance accuracy disappears.}
A model retrained on a subsample in which \texttt{sex} no longer correlates with the outcome performs close to chance ($53.3 \pm 0.5\%$). 

\textbf{On the brain-age model, ICON finds no substantial shortcut among the provided concepts, whereas the probe surfaces many concepts (\autoref{fig:neuro}b).}
Our brain-age model, a Simple Fully Convolutional Network (SFCN)~\cite{peng19_brainage}, predicts age with a mean absolute error (MAE) of $2.89 \pm 0.02$ years ($R^2 = 0.750 \pm 0.004$).
At $\mathbf{H}_L$, ICON attributes $\approx77\%$ of the variance to the outcome, \texttt{age}, while a few other concepts, such as the scan's \texttt{acquisition\_date}, together explain under $1\%$ of the variance (\autoref{fig:neuro}b, left; \autoref{tab:neuro_ood}).
Unlike ICON, probes provide no common scale for comparing concepts, so their scores can only be read as a ranking.
Probes rank many concepts highly along with \texttt{age} (\autoref{fig:neuro}b, right), several of which correlate with age (Extended Data \autoref{fig:ED:neuro_corr})~\cite{hyatt20_covarreview, alfaro21_ukbbconfound}.
ICON also flags an inadequate concept set by showing that $\approx22\%$ of the variance in the scalar prediction $\mathbf{H}_L$ cannot be linearly explained by the $30$ provided concepts or the outcome.
Neither probes nor CAVs provide such a statement of concept-set adequacy.

\textbf{Two independent tests support ICON's reading of the brain-age model rather than the probe's (\autoref{tab:neuro_ood}).}
We take the five concepts other than \texttt{age} that each method ranks highest at $\mathbf{H}_L$, which gives eight concepts in \autoref{tab:neuro_ood}.
Using two complementary tests, out-of-distribution (OOD) validation and a conditional independence test (CIT), we check which of these concepts the prediction $\mathbf{H}_L$ depends on beyond \texttt{age} (Methods~\ref{sec:methods:exp3}).
For six of the eight concepts we could construct an OOD test set, and on none of them does the prediction error differ significantly from that on a matched control set (mean $|\Delta\text{MAE}| \le 0.069$ years).
This is consistent with ICON's reading that none of these concepts is used to an appreciable degree ($\le1\%$) compared with the outcome ($77\%$).
The more sensitive CIT detects small dependencies between the prediction and three concepts, which are exactly the three that ICON ranks highest after the outcome. This suggests that even ICON's small scores reflect real dependencies.
The three concepts ranked highly only by the probes are flagged by neither of the two validations.

\textbf{ICON can also compare architectures and training stages (Extended Data \autoref{fig:neuro_diagnostics}).}
In the brain-age model, scanner-position concepts dominate the small explained share in the early training epochs and give way to \texttt{age} as training converges. ICON shows at which stage, and in which layers, the representation stops being dominated by acquisition factors (Extended Data \autoref{fig:neuro_diagnostics}a).
Training a different architecture for the binge-drinking task (SFCN) reproduces the dominance of \texttt{sex}, suggesting that the shortcut originates in the data rather than in the architecture (Extended Data \autoref{fig:neuro_diagnostics}b).

\begin{table}[!htbp]
\centering
\begin{minipage}{\linewidth}
\centering
\caption{\textbf{In the brain-age model, concepts flagged only by the linear probe are not supported by either of the two independent validations.}
We take the five concepts that each method ranks highest at $\mathbf{H}_L$, which gives eight concepts.
ICON importance and probe score at $\mathbf{H}_L$ for the eight selected concepts (means over the five cross-validation folds), with their rankings shown in parentheses. 
A dash in a rank column means the concept falls outside that method's top five.
$\Delta$MAE denotes the change in mean absolute error between the out-of-distribution (OOD) and matched control sets in years (mean $\pm$ s.d.) and `infeasible' means an OOD set could not be constructed for that concept (Methods~\ref{sec:methods:exp3}). 
The final column shows the mean $\chi^2$ of a conditional independence test (CIT) of the prediction $\mathbf{H}_L$ against the concept given \texttt{age} and the remaining seven concepts (largest $P$ across folds in parentheses, Methods~\ref{sec:methods:exp3}). The CIT uses the $n = 3{,}738$ auditing-dataset participants with annotations for all eight concepts.
\textbf{$P$ values:} Because the folds reuse the same participants, a concept counts as significant only if its largest $P$ value across the folds clears a Bonferroni-corrected level, $\alpha = 0.05/6$ for $\Delta$MAE and $\alpha = 0.05/8$ for the CIT. 
No $\Delta$MAE is significant and the smallest $P$ value in any fold is $0.07$. In the CIT, asterisks mark the three significant concepts which match with the top three concepts of ICON after \texttt{age}.
}
\label{tab:neuro_ood}
\setlength{\tabcolsep}{4pt}
\begin{tabular}{lcccc}
\toprule
\input{figures/neuro_results-ood.tex}
\end{tabular}
\end{minipage}
\end{table}

%% file: figures/neuro_results-ood.tex
Concept & \shortstack{ICON\\score (rank)} & \shortstack{Probe\\score (rank)} & \shortstack{Independent validation 1:\\$\Delta$MAE (years)} & \shortstack{Independent validation 2:\\CIT: $\chi^2$ (worst-case $P$)} \\
\midrule
\texttt{acquisition date} & 0.004 (1) & 0.001 (--) & infeasible & $89.30$ ($<0.001$)\textsuperscript{*} \\
\texttt{snap reaction time} & 0.001 (2) & 0.082 (3) & $+0.010 \pm 0.022$ & $19.15$ ($<0.001$)\textsuperscript{*} \\
\texttt{alcohol use frequency} & 0.001 (3) & 0.007 (--) & $-0.031 \pm 0.016$ & $28.99$ ($<0.001$)\textsuperscript{*} \\
\texttt{scanner table position} & 0.001 (4) & 0.000 (--) & $-0.069 \pm 0.029$ & $1.87$ ($0.48$) \\
\texttt{systolic blood pressure} & 0.001 (5) & 0.072 (5) & $+0.038 \pm 0.043$ & $10.89$ ($0.0065$) \\
\texttt{household income} & 0.000 (--) & 0.072 (4) & $+0.024 \pm 0.043$ & $8.06$ ($0.011$) \\
\texttt{sex} & 0.000 (--) & 0.108 (2) & $-0.002 \pm 0.025$ & $2.40$ ($0.44$) \\
\texttt{employment status} & 0.000 (--) & 0.360 (1) & infeasible & $6.41$ ($0.15$) \\

%% file: sections/discussion.tex
\section{Discussion}\label{Discussion}

C-XAI methods such as linear probes and CAVs have largely re-derived variable importance without drawing on the statistics literature that has studied this problem for decades~\cite{gromping15_statsvarimp}. 
ICON adapts this literature and reframes C-XAI as a multivariate, layer-wise variance-decomposition problem.
ICON preserves the flexibility of probes and CAVs, allowing us to interpret any trained model with any user-defined concept annotated in the auditing dataset.
At the same time, ICON produces fewer false positives than these methods (\autoref{fig:res-toybrains}c,d, \autoref{fig:isic_results}c,d), makes concept importance scores comparable across layers and concept types, and quantifies how much of a representation the provided concepts leave unexplained.

These properties translate into three practical benefits.
First, ICON's low false-positive rate yields explanations that are sparse, leaving only a few concepts that carry appreciable importance (\autoref{fig:res-toybrains}d), as seen in the neuroimaging experiments (\autoref{fig:neuro}). Such sparse attributions can serve as actionable hypotheses for follow-up experiments.
Independent validations supported the small number of concepts that ICON highlighted but found no evidence for many of the concepts that the probe ranked highly (\autoref{tab:neuro_ood}). 
Second, layer-wise comparability lets us follow the trajectory of concepts from the input representation to the prediction, enabling us to distinguish between concepts that are merely encoded and those that drive the prediction.
For example, in the skin-cancer experiment, the inserted \texttt{microscope} artifact is encoded in the early layers of both models but survives into the prediction layers only in the shortcut-affected one. In the binge-drinking model, the same trajectory flags \texttt{sex} as a shortcut.
Finally, a large unexplained share signals when the provided concepts are inadequate to linearly explain a representation (Extended Data \autoref{fig:isic_results_abs}a, \autoref{fig:neuro}b). This helps guard against overconfidence in the provided concept set and the linearity assumption.
 
When using ICON in practice, we recommend providing ICON with all available annotated concepts, including the outcome, since a larger set raises the chance of including the relevant concepts. Our simulations show that ICON's accuracy does not degrade as the number of provided concepts increases, unlike the univariate baselines (Extended Data \autoref{fig:ED:toybrains-pqsweep}a).
We further recommend applying ICON to multiple layers, reading the trajectory of concept importance rather than any single layer, and checking the unexplained share.
Because ICON is observational, the concepts it nominates should be validated with targeted tests such as out-of-distribution testing or conditional independence tests~\cite{shah20_GCM, simnacher26_DNCIT}. 

Three limitations qualify ICON's claims.
First, like linear probes and CAVs, ICON assumes linear encoding, so a nonlinearly encoded concept may be under-attributed, with its nonlinear part appearing in the unexplained share.
Second, the unexplained share mixes two sources, concepts missing from the provided set and nonlinear encoding, which we do not yet separate.
Finally, like the baselines, ICON cannot discover concepts outside the provided set. The unexplained share indicates that something is missing, but not what.

Future work could test whether steering models along ICON's directions reduces the collateral problems caused by multicollinearities, reported in CAV-based steering studies~\cite{zhang25_CAV_LLMsteering, erogullari25_disentangleCAVs}. Furthermore, future works can relax the linear-encoding assumption by testing nonlinear estimators such as kernel PLS.
Finally, the conditions under which ICON's observational decomposition can be read causally also remain open for investigation.
Nonetheless, ICON resolves a fundamental limitation of concept-based explainability: because existing decodability-based methods ask the wrong question, they cannot distinguish concepts that merely reflect multicollinearity in the auditing dataset from concepts that influence model predictions.
When ICON is used to audit a model for shortcuts or to generate hypotheses, it requires only trained-model activations and a few hundred annotated samples for a stable decomposition (Extended Data \autoref{fig:ED:toybrains-pqsweep}b), making it practical for biomedical cohorts as DNNs become increasingly prevalent in biomedical research and clinical practice.
More broadly, ICON enables comparisons of concept composition across training stages and model architectures (Extended Data \autoref{fig:neuro_diagnostics}), opening new avenues for model inspection.

%% file: sections/methods.tex

\section{Methods}\label{methods}

\subsection{ICON Decomposition}\label{sec:methods:math}

\textbf{Formal motivation and definition:} 
Let us consider a DNN model trained on a dataset $\{\mathbf{X}, \mathbf{y} \}$ where $\mathbf{X}$ denotes the input features (e.g., dermoscopic images or brain MRI scans) and $\mathbf{y}$ denotes the outcome variable being predicted (e.g., melanoma or a brain disorder). A DNN model with $L$ layers can be expressed as a series of transformations $M: \mathbf{X} \xrightarrow{M_1} \mathbf{H}_1 \xrightarrow{M_{2}} \cdots{} \mathbf{H}_{L-2} \xrightarrow{M_{L-1}} \mathbf{H}_{L-1} \xrightarrow{M_L} \mathbf{H}_L$, whereby $\mathbf{H}_l \in \mathbb{R}^{n \times p_l}$ represents the latent representation learned by the model at an intermediary layer $l$. 
In C-XAI, the goal is to interpret $\mathbf{H}_l$ using a set of concept variables $\mathbf{C} = [\mathbf{c}_1, \ldots, \mathbf{c}_q]$ by estimating the importance $\mathrm{Imp}(\mathbf{c}_k)$ at layer $l$. 
We call the $n$ held-out samples annotated with the concepts in $\mathbf{C}$ and the outcome $\mathbf{y}$ the \emph{auditing dataset}, on which the C-XAI importances are computed. The auditing dataset is disjoint from the training set, which is used for model optimisation.

Current C-XAI methods such as linear probes and CAVs estimate the importance of each concept individually (univariate estimation of $\mathrm{Imp}(\mathbf{c}_k)$). It is often the case that $\mathbf{c}_k$ is correlated with another important concept $\mathbf{c}_j$ or the outcome variable $\mathbf{y}$. Estimating $\mathrm{Imp}(\mathbf{c}_k)$ while omitting the other correlated variables therefore conflates the importance of $\mathbf{c}_k$ with theirs.
In a classical regression model, this is the omitted-variable bias of a coefficient when a correlated predictor is left out of the model \cite[Ch.~10]{kutner05_statsbook}.
Throughout this paper we call a concept a \emph{false positive} when a C-XAI method assigns it appreciable importance although the model's representation does not depend on it.

In ICON decomposition, we reduce this bias by controlling for all the provided concept variables and for the outcome jointly (i.e., multivariate estimation of $\mathrm{Imp}(\mathbf{c}_k)$).
We collect them into a single matrix
\begin{equation}\label{eq:cprime}
\mathbf{C}' = [\mathbf{c}_1, \mathbf{c}_2, \ldots, \mathbf{c}_q, \mathbf{y}] \in \mathbb{R}^{n \times (q+1)},
\end{equation}
and estimate $\mathrm{Imp}(\mathbf{c}_k)$ for all variables in $\mathbf{C}'$ jointly, so that each concept is credited only with variance in $\mathbf{H}_l$ that is not attributable to a more strongly contributing variable in $\mathbf{C}'$ (Step 2b).
Including $\mathbf{y}$ in $\mathbf{C}'$ lets the outcome compete with the concepts for shared variance, so a concept that is merely a correlate of the outcome receives little or no importance whenever the outcome contributes more strongly. 

\textbf{Derivation}: 
ICON decomposition consists of two steps. First, we use PLS to capture latent associations between the concepts $\mathbf{C}'$ and the representation $\mathbf{H}_l$, and then we allocate the explained variance among the variables in $\mathbf{C}'$ using Type~I sums of squares. Before applying PLS we standardise every variable in $\mathbf{C}'$ and every column of the latent representation $\mathbf{H}_l \in \mathbb{R}^{n \times p_l}$ to zero mean and unit standard deviation using the auditing dataset statistics. The same preprocessing is also applied to all compared methods.

\emph{Step (1) Use PLS to capture latent associations}: 
PLS is a multivariate statistical method for finding linear associations between two sets of variables when both sets are high-dimensional and internally collinear, which is exactly the situation we face with the DNN layer representation $\mathbf{H}_l$ and the concept matrix $\mathbf{C}'$.

At each iteration $i$, the optimisation finds weight vectors $\mathbf{u}_{l,i} \in \mathbb{R}^{p_l}$ and $\mathbf{v}_{l,i} \in \mathbb{R}^{q+1}$ that maximise:
\begin{equation}
\begin{aligned}
\max_{\mathbf{u}_{l,i},\, \mathbf{v}_{l,i}} \;\text{Cov}\!\left(\mathbf{H}_l^{(i)} \mathbf{u}_{l,i},\; \mathbf{C}'^{(i)} \mathbf{v}_{l,i}\right) \;=\; \tfrac{1}{n-1}\max_{\mathbf{u}_{l,i},\, \mathbf{v}_{l,i}} \;\mathbf{u}_{l,i}^\top {\mathbf{H}_l^{(i)}}^\top \mathbf{C}'^{(i)} \mathbf{v}_{l,i} \\ 
\text{where $\tfrac{1}{n-1}{\mathbf{H}_l^{(i)}}^\top\mathbf{C}'^{(i)}$ is the cross-covariance matrix,}\\
\text{subject to } \|\mathbf{u}_{l,i}\|_2 = \|\mathbf{v}_{l,i}\|_2 = 1. 
\end{aligned}
\label{eq:icon_cross_cov}
\end{equation}
This optimisation yields a latent concept $\tilde{\mathbf{c}}_{l,i} = \mathbf{C}'^{(i)} \mathbf{v}_{l,i}$, called the Independent Canonical cONcept (ICON), and a corresponding latent activation vector $\tilde{\mathbf{h}}_{l,i} = \mathbf{H}_l^{(i)} \mathbf{u}_{l,i}$, called the ICON activation vector. The pair $\{\tilde{\mathbf{c}}_{l,i}, \tilde{\mathbf{h}}_{l,i}\}$ captures the largest covariance attainable between $\{\mathbf{C}'^{(i)}, \mathbf{H}_l^{(i)}\}$.  
The first ICON pair $\{\tilde{\mathbf{c}}_{l,1}, \tilde{\mathbf{h}}_{l,1}\}$ is computed directly on the original matrices ($\mathbf{H}_l^{(1)} = \mathbf{H}_l$ and $\mathbf{C}'^{(1)} = \mathbf{C}'$), but for all subsequent iterations, the optimisation is performed on the residuals:
\begin{equation}
\mathbf{H}_l^{(i+1)} = \mathbf{H}_l^{(i)} - \tilde{\mathbf{h}}_{l,i} \, \boldsymbol{\gamma}_{l,i}^\top, \quad
\mathbf{C}'^{(i+1)} = \mathbf{C}'^{(i)} - \tilde{\mathbf{c}}_{l,i} \, \boldsymbol{\delta}_{l,i}^\top
\label{eq:pls_deflation}
\end{equation}
where the loadings $\boldsymbol{\gamma}_{l,i} \in \mathbb{R}^{p_l}$ and $\boldsymbol{\delta}_{l,i} \in \mathbb{R}^{q+1}$ are the least-squares projections of each block onto its own score, $\boldsymbol{\gamma}_{l,i} = {\mathbf{H}_l^{(i)}}^\top \tilde{\mathbf{h}}_{l,i} / (\tilde{\mathbf{h}}_{l,i}^\top \tilde{\mathbf{h}}_{l,i})$ and $\boldsymbol{\delta}_{l,i} = {\mathbf{C}'^{(i)}}^\top \tilde{\mathbf{c}}_{l,i} / (\tilde{\mathbf{c}}_{l,i}^\top \tilde{\mathbf{c}}_{l,i})$. In practice, we add pairs one at a time and stop when a new pair explains less than $1\%$ of the total squared cross-covariance~\cite{wold2001_plschemo}. 

Since each block is deflated by its own score, all ICON pairs are mutually orthogonal: $\tilde{\mathbf{h}}_{l,i}^\top \tilde{\mathbf{h}}_{l,j} = 0$ and $\tilde{\mathbf{c}}_{l,i}^\top \tilde{\mathbf{c}}_{l,j} = 0$ for $i \neq j$.
In the end, this produces a set of $m$ orthogonal ICON activations $\{\tilde{\mathbf{h}}_{l,1}, \tilde{\mathbf{h}}_{l,2}, \ldots, \tilde{\mathbf{h}}_{l,m}\}$ that span a low-rank subspace of $\mathbf{H}_l$~\cite{wegelin00_plssklearn}.
From a CAV practitioner's perspective, the weight vector $\mathbf{u}_{l,i}$ is the CAV of the latent concept $\tilde{\mathbf{c}}_{l,i}$.

\emph{Step (2) Deriving concept importance by decomposing the variance explained}:
We define the importance of each concept $\mathrm{Imp}(\mathbf{c}_k), \; \forall \mathbf{c}_k \in \mathbf{C}'$, as the share of total variance explained by $\mathbf{c}_k$ in $\mathbf{H}_l$. To obtain this, we first derive the variance explained by the orthogonal latent concepts $\tilde{\mathbf{c}}_{l,i}$ and then decompose this among the original concept and outcome variables using Type~I sums of squares.

\emph{(2a) Deriving variance explained by the latent concepts $\tilde{\mathbf{c}}_{l,i}$}:
For each ICON pair, we can derive the variance explained by the latent concept $\tilde{\mathbf{c}}_{l,i}$ in $\mathbf{H}_l$ as follows,
\begin{align*}
&\text{The PLS decomposes the representation as $\mathbf{H}_l = \sum_{i=1}^{m} \tilde{\mathbf{h}}_{l,i} \, \boldsymbol{\gamma}_{l,i}^\top + \mathbf{E}_l$,}\\
&\text{where $\boldsymbol{\gamma}_{l,i}$ are the loadings at iteration $i$ and $\mathbf{E}_l$ is the residual after repeated}\\ &\text{subtraction in \autoref{eq:pls_deflation}. Applying the variance operator to both sides,}\\
    &\text{Var}(\mathbf{H}_l)
        = \sum_{i=1}^{m} \|\boldsymbol{\gamma}_{l,i}\|^2\,\text{Var}(\tilde{\mathbf{h}}_{l,i} )  + \text{Var}(\mathbf{E}_l)\text{.} \\ 
    &\text{Here, the cross-covariance terms vanish since }\tilde{\mathbf{h}}_{l,i}^\top \tilde{\mathbf{h}}_{l,j} = 0 \text{ for } i \neq j \text{, by design.}\\ 
    &\text{Regressing the ICON activation on the concepts ($\tilde{\mathbf{h}}_{l,i} \sim \mathbf{C}'$) using OLS}\\ 
    &\text{yields $\tilde{\mathbf{h}}_{l,i} = \mathbf{C}'\mathbf{b}_{i} + \epsilon'_i$. So its variance breaks down as}\\ &\text{Var($\tilde{\mathbf{h}}_{l,i}) = \text{Var}(\mathbf{C}'\mathbf{b}_{i})  + \text{Var}(\epsilon'_i)$ since Cov($\mathbf{C}'\mathbf{b}_{i},\, \epsilon'_i) = 0$ by construction.}\\ 
    &\text{If we denote the share of $\text{Var}(\tilde{\mathbf{h}}_{l,i})$ explained by $\mathbf{C}'$ as $R^2_{l,i}$, then}\\
    &\text{Var($\tilde{\mathbf{h}}_{l,i}) = \text{Var}(\tilde{\mathbf{h}}_{l,i})\,R^2_{l,i}  + \text{Var}(\epsilon'_i)$. Substituting this gives}\\
    &\text{Var}(\mathbf{H}_l)
        = \sum_{i=1}^{m} \|\boldsymbol{\gamma}_{l,i}\|^2\,\text{Var}(\tilde{\mathbf{h}}_{l,i})\,R^2_{l,i}  +
            \sum_{i=1}^{m} \|\boldsymbol{\gamma}_{l,i}\|^2\,\text{Var}(\epsilon'_i) + \text{Var}(\mathbf{E}_l)\text{.}\\
    &\text{Dividing both sides by $\text{Var}(\mathbf{H}_l)$,}\\
        &1 = \sum_{i=1}^{m} \underbrace{\frac{\|\boldsymbol{\gamma}_{l,i}\|^2\,\text{Var}(\tilde{\mathbf{h}}_{l,i})\,R^2_{l,i}}{\text{Var}(\mathbf{H}_l)}}_{\mathrm{Imp}(\tilde{\mathbf{c}}_{l,i})}  +
            \underbrace{\frac{\sum_{i=1}^{m}\|\boldsymbol{\gamma}_{l,i}\|^2\,\text{Var}(\epsilon'_i) + \text{Var}(\mathbf{E}_l)}{\text{Var}(\mathbf{H}_l)}}_{\sigma^2_l \text{ (residual variance)}}
\end{align*}
Thus, we obtain a clean decomposition by defining the importance of each ICON as the ratio: 
\begin{equation}
\mathrm{Imp}(\tilde{\mathbf{c}}_{l,i}) = \frac{\|\boldsymbol{\gamma}_{l,i}\|^2\,\text{Var}(\tilde{\mathbf{h}}_{l,i})\,R^2_{l,i}}{\text{Var}(\mathbf{H}_l)}, \quad\text{such that}\quad \sum_{i=1}^{m}\mathrm{Imp}(\tilde{\mathbf{c}}_{l,i}) + \sigma^2_l = 1
\label{eq:icon_step1}
\end{equation}

The residual variance $\sigma^2_l$ is the share of the representation's variance that cannot be assigned to any concept or to the outcome variable in $\mathbf{C}'$ under the linearity assumption. 

\emph{(2b) Decomposing $\mathrm{Imp}(\tilde{\mathbf{c}}_{l,i})$ into the importance of the original concepts in $\mathbf{C}'$}:

\emph{Tackling the non-identifiability challenge.} Although the ICONs are linearly independent by construction, the original variables forming each ICON,
$\tilde{\mathbf{c}}_{l,i} = v_{l,i,1}\, \mathbf{c}_1 + v_{l,i,2}\, \mathbf{c}_2 + \cdots + v_{l,i,q}\, \mathbf{c}_q + v_{l,i,q+1}\, \mathbf{y}$,
can be multicollinear. 
Multicollinear variables share overlapping variance that cannot be uniquely attributed to any one of them, a fundamental limitation known as non-identifiability~\cite[Ch.~10]{kutner05_statsbook}. 
Because shared variance is not uniquely identifiable, we adopt an explicit allocation convention suited to our screening objective~\cite{gromping15_statsvarimp}.
Specifically, we order variables within each ICON by their squared weights $v^2_{l,i,k}$ and then use sequential (Type~I) sums of squares. 
This convention allocates the shared part of $\mathrm{Imp}(\tilde{\mathbf{c}}_{l,i})$ to the dominant variables, that is, those with the largest squared weights $v^2_{l,i,k}$, which absorb any variance they share with weaker contributors; this reduces false positives during screening.

\emph{Type~I sum of squares.} Let $\pi_i$ denote the ordering of the variables that compose an ICON $\tilde{\mathbf{c}}_{l,i}$, such that $v^2_{l,i,\pi_i(1)} \geq v^2_{l,i,\pi_i(2)} \geq \cdots \geq v^2_{l,i,\pi_i(q+1)}$. 
We sequentially regress the ICON activation vector $\tilde{\mathbf{h}}_{l,i}$ on the original variables in $\mathbf{C}'$, one by one, following this order, which yields a set of 
incremental sums of squares $\{\Delta\text{SS}_{i,\pi_i(1)}, \ldots, \Delta\text{SS}_{i,\pi_i(q+1)}\}$, where $\Delta\text{SS}_{i,\pi_i(j)}$ is the additional variance in $\tilde{\mathbf{h}}_{l,i}$ explained by $\mathbf{c}_{\pi_i(j)}$ after accounting for all its higher-ranked variables $\mathbf{c}_{\pi_i(1)}, \ldots, \mathbf{c}_{\pi_i(j-1)}$. We write $\Delta\text{SS}_{i,k}$ for the increment belonging to variable $\mathbf{c}_k$, that is, $\Delta\text{SS}_{i,\pi_i(j)}$ with $\pi_i(j) = k$. 

Therefore, the importance of concept $\mathbf{c}_k$ within ICON $\tilde{\mathbf{c}}_{l,i}$ is
\begin{equation*}
\mathrm{Imp}(\mathbf{c}_k,\, \tilde{\mathbf{c}}_{l,i}) 
= \frac{\Delta\text{SS}_{i,k}}{\sum_{j=1}^{q+1} \Delta\text{SS}_{i, \pi_i(j)}}
\mathrm{Imp}(\tilde{\mathbf{c}}_{l,i}) \text{.}
\end{equation*}
The incremental sums of squares add up to the variance in $\tilde{\mathbf{h}}_{l,i}$ that $\mathbf{C}'$ explains, so $\sum_{j}\Delta\text{SS}_{i,\pi_i(j)} = \text{Var}(\tilde{\mathbf{h}}_{l,i})\,R^2_{l,i}$. Substituting this and \autoref{eq:icon_step1},
\begin{equation}
\mathrm{Imp}(\mathbf{c}_k,\, \tilde{\mathbf{c}}_{l,i}) = \frac{\|\boldsymbol{\gamma}_{l,i}\|^2\,\Delta\text{SS}_{i,k}}{\text{Var}(\mathbf{H}_l)} \text{, such that } \textstyle\sum_{k}\mathrm{Imp}(\mathbf{c}_k,\, \tilde{\mathbf{c}}_{l,i}) = \mathrm{Imp}(\tilde{\mathbf{c}}_{l,i})\text{.}
\label{eq:icon_step2}
\end{equation}

The total importance of a concept $\mathbf{c}_k$ across all ICONs is then:
\begin{equation}
\mathrm{Imp}(\mathbf{c}_k) = \sum_{i=1}^{m} \mathrm{Imp}(\mathbf{c}_k,\, \tilde{\mathbf{c}}_{l,i})
\label{eq:total_concept_importance}
\end{equation}
We allocate by $\Delta\text{SS}$ rather than by $v^2_{l,i,k}$ directly because the $v^2_{l,i,k}$ weights describe how the ICON was built out of $\mathbf{C}'$, whereas the quantity we want to allocate is variance in $\tilde{\mathbf{h}}_{l,i}$. 

\textbf{Beneficial properties of ICON decomposition:}
This formulation gives the ICON importance scores four properties that directly tackle the three limitations of existing C-XAI metrics we discuss in the Introduction:
\begin{enumerate}
    \item \textbf{Bounded and sum-constrained:} $\mathrm{Imp}(\mathbf{c}_k) \in [0,1]$ and $\sum_{k} \mathrm{Imp}(\mathbf{c}_k) + \sigma^2_l = 1$, enabling direct comparison across concept types and layers.
    \item \textbf{Transparent about unexplained variance:} $\sigma^2_l$ quantifies how much of the latent representation lies outside the provided concept set and linear assumptions, guarding against overconfidence in an incomplete concept set~\cite{poeta23_ConceptXAIsurvey}, or over-reliance on the linearity assumption~\cite{crabbe22_CAR}.
    \item \textbf{Transparent about entangled concepts:} Within each ICON, the squared weights $v^2_{l,i,k}$ from Step~1 record which concepts share variance, making multicollinear groups explicit to the practitioner.
    \item \textbf{Down-weights weaker, correlated concepts:} The dominant variable within each ICON absorbs the variance it shares with weaker correlated ones. This strategy reduces false positives by revealing the variables that drive shortcut learning rather than spuriously correlated concepts, as we demonstrate in the experiments.  
\end{enumerate}

\subsection{Visualising layer-wise ICON decompositions as a trajectory of concept importance}\label{sec:methods:viz}
We visualise the ICON concept importance at each layer as stacked horizontal bars that represent the variance partition of that layer. We then stack the decompositions of all layers, ordered from the input layer at the bottom to the output prediction at the top (for example, \autoref{fig:isic_qualitative} and \autoref{fig:neuro}).
The lengths of the horizontal bars sum to one at each layer (\autoref{eq:icon_step1}), and the fraction of representation variance not explained by any provided concept or the outcome, $\sigma^2_l$, is drawn as a separate unexplained segment.
Although several concepts can contribute to each bar, only the variables that have an importance $\ge5\%$ at at least one layer are named in the legend.

\subsection{Experiment setup}\label{sec:methods:setup}

\textbf{Comparing with baseline methods.}\label{sec:methods:baselines} Across three experiments, we compare ICON against two groups of baseline methods.
The first group consists of the following established C-XAI methods:

\begin{enumerate}
\item \emph{Linear probes}~\cite{alain16_probing}: For each $\mathbf{c}_k \in \mathbf{C}'$, we fit a logistic (categorical) or linear (continuous) regression predicting $\mathbf{c}_k$ from $\mathbf{H}_l$. We perform three-fold cross-validation and use the mean test fit quality as the importance score (cross-validated $R^2 \in [0,1]$ for continuous, Cohen's $\kappa \in [0,1]$ for categorical with negative values truncated to $0$; $0$ denotes chance-level prediction). 
Linear probes are the most widely used C-XAI method, e.g., to assess representation quality in self-supervised models~\cite{chen20_SimCLR} or detect linguistic structure in language models~\cite{hewitt19_probinglinearbetter}. 

\item \emph{CAV-classic}~\cite{kim18_TCAV}: A concept activation vector (CAV) denotes a direction in the layer's activation space, obtained from a learned linear probe. 
Testing with CAVs (TCAV)~\cite{kim18_TCAV} measures how much the model's output changes when the layer's activations are perturbed along the CAV direction. 
In CAV-classic, we use this TCAV sensitivity metric as the importance score. 
Raw TCAV scores equal $0.5$ when a concept has no directional effect. Following Pahde et al.~\cite{pahde24_signalCAV}, we renormalise the TCAV score as $|q_{\text{CAV}} - 0.5| / 0.5$, to map CAV-classic to a $[0,1]$ scale, comparable with all other methods.  
CAV-based approaches are widely used to detect shortcuts in biomedical applications~\cite{pahde25_thesis} and to interpret language models~\cite{zhang25_CAV_LLMsteering}.

\item \emph{CAV-signal}~\cite{pahde24_signalCAV}: Rather than fitting a probe to obtain the CAV direction, CAV-signal defines the CAV as the difference between the mean activation in $\mathbf{H}_l$ when $\mathbf{c}_k = 1$ and when $\mathbf{c}_k = 0$ (defined for binary $\mathbf{c}_k$). Pahde et al.~\cite{pahde24_signalCAV} argue that this simpler signal direction makes CAVs less susceptible to confounding by correlated concepts encoded in $\mathbf{H}_l$. As with CAV-classic, the importance score is the symmetric magnitude $|q_{\text{CAV}} - 0.5| / 0.5$ of the directional-derivative statistic, computed along this signal direction.

\end{enumerate}

The second group consists of classical statistical methods for variable importance in regression (see Gr\"omping~\cite{gromping15_statsvarimp} for a review), which we consider as potential alternatives to ICON's PLS-based variance decomposition. 
\begin{enumerate} \setcounter{enumi}{3}

\item \emph{Marginal $R^2$ importance}: The $R^2$ of a univariate regression of $\mathbf{H}_l$ on $\mathbf{c}_k$ alone, ignoring all other concepts. This captures the marginal association between $\mathbf{c}_k$ and $\mathbf{H}_l$. However, when concepts are correlated, their shared variance is attributed to each concept individually, inflating importance scores.

\item \emph{Standardised regression coefficients}: We fit a multi-output linear regression,
\begin{equation}\label{eq:fullmodel}
\mathbf{H}_l = \mathbf{C}'\mathbf{W} + \mathbf{E}, \quad \mathbf{W} \in \mathbb{R}^{(q+1) \times p_l},\; \mathbf{E} \in \mathbb{R}^{n \times p_l},
\end{equation}
which decomposes into $p_l$ linear regressions, one per output dimension. We use the mean squared coefficient $\frac{1}{p_l}\sum_{j=1}^{p_l} W_{k,j}^{2}$ as the importance of concept $\mathbf{c}_k$. This accounts for the presence of other concepts, but estimated coefficients are known to be numerically unstable under multicollinearity~\cite{gromping15_statsvarimp}.

\item \emph{Partial $R^2$ importance}: The drop in $R^2$ when $\mathbf{c}_k$ is removed from the full model (\autoref{eq:fullmodel}), i.e., $R^2(\mathbf{W}) - R^2(\mathbf{W}_{\setminus k})$, where $\mathbf{W}_{\setminus k}$ is the model refitted without $\mathbf{c}_k$, and $R^2$ is the coefficient of determination averaged across the $p_l$ output dimensions. This corresponds to Type~III (``last'') sum of squares~\cite{gromping15_statsvarimp}, quantifying the unique variance explained by $\mathbf{c}_k$ after accounting for all other concepts. Under strong multicollinearity, this unique contribution may shrink to near zero even for genuinely important concepts. 
\item \emph{Shapley $R^2$ importance}: The Shapley value~\cite{shapley1953value} of $\mathbf{c}_k$ in the linear regression model from \autoref{eq:fullmodel} is the average marginal contribution of $\mathbf{c}_k$ across all possible orderings in which it could enter a Type~I sum-of-squares decomposition. Unlike all above methods, Shapley values explicitly ensure that any shared variance among correlated concepts is distributed symmetrically~\cite{shapley1953value, gromping15_statsvarimp}. We estimate it from the multivariate $R^2$ in \autoref{eq:fullmodel} using the Lindeman, Merenda, and Gold method~\cite{gromping15_statsvarimp}.
\end{enumerate}

\textbf{Normalising concept importance to enable comparison.} 
Importance scores from different methods, such as ICON, CAVs and linear probes, have different natural scales and the literature does not provide a unified notion that makes these importance scores directly comparable~\cite{poeta23_ConceptXAIsurvey}.  
We place all the methods on a common simplex by normalising their importance scores over the supplied variable set $\mathbf{C}'$:
\begin{equation}\label{eq:relimp}
\bar{\mathrm{Imp}}(\mathbf{c}_k) = \frac{\mathrm{Imp}(\mathbf{c}_k)}{\sum_{j=1}^{q+1} \mathrm{Imp}(\mathbf{c}_j)}.
\end{equation}
This places all methods on the same fixed budget and reduces the effect of layer dimensionality on the raw scores. 
We also report the unnormalised scores alongside the normalised ones wherever a conclusion could depend on this choice (e.g., Extended Data \autoref{fig:isic_results_abs}).

\subsection{Statistics and reproducibility}\label{sec:methods:stats}
Unless stated otherwise, summary values are reported as mean $\pm$ s.d. The meaning of $n$ depends on the analysis and is specified in the corresponding results and figure captions.
For Experiment 1, we report the median and the 25th--75th percentile range across the 90 models because the error distribution is skewed.
Experiments 1 and 2 are descriptive: C-XAI methods are compared against a known ground truth, and no hypothesis tests are performed.
All hypothesis tests in Experiment 3 are two-sided. Because the five folds reuse participants, we do not pool their $P$ values. Instead, an effect is considered significant only if the largest $P$ value across the five folds falls below the Bonferroni-adjusted significance threshold: $\alpha = 0.05/6$ for the six $\Delta$MAE tests and $\alpha = 0.05/8$ for the eight conditional independence tests.

No formal sample-size calculation was performed as cohort sizes were determined by the availability of concept annotations in UK Biobank and by the size of the ISIC 2019 release. Experiment-specific exclusions are reported in the corresponding experiment descriptions. No additional data were excluded.
For procedures that use random seeds, the seeds are fixed in the released code, including artifact insertion and the dataset splits in Experiment 2. The ISIC models and the reported UK Biobank models were trained without fixed random seeds. The code archive~\cite{rane26_iconcode} contains the aggregate results needed to regenerate \autoref{fig:neuro} and \autoref{tab:neuro_ood}.


\subsection{Experiment 1: Simulation experiments with ToyBrains}\label{sec:methods:exp1}

\textbf{ToyBrains dataset configuration.}
ToyBrains~\cite{roshanrane_toybrains_2024} generates $64\times64$ RGB images ($\mathbf{X}$) whose visual attributes are specified by sixteen generative variables $\mathbf{L} = \{\mathbf{l}_1,\dots,\mathbf{l}_{16}\}$ (\autoref{fig:res-toybrains}a). These sixteen generative image attributes $\mathbf{L}$ are considered unobservable to C-XAI methods and all paths into the image run through them. We also generate a binary outcome variable $\mathbf{y}$ and a pool of $100$ concepts $\mathbf{C}=\{\mathbf{c}_1, \dots, \mathbf{c}_{100}\}$.
Together, these variables $\{\mathbf{L}, \mathbf{C}, \mathbf{y}\}$ form the nodes of the data-generating process and their causal coefficients form the edges. 
Every variable is discrete in this directed acyclic graph. In the absence of any incoming edge, the variable is uniformly distributed over its states, and an edge coefficient acts by shifting the probability of the target variable through a logistic link. For example, if the target variable $\mathbf{z}$ is binary then $\beta = 0$ has no effect as it keeps $P(\mathbf{z}=1)=50\%$ as before, while $\beta = 2.20$ sets $P(\mathbf{z}=1)=90\%$. Variables with a fixed number of ordinal states are used to emulate continuous variables.

For this experiment, we design a graph in which the image and the outcome are related through one `true' signal pathway and two confounder-driven pathways (\autoref{fig:res-toybrains}b).
Each attribute in $\mathbf{L}$ is uniquely assigned: a specific attribute, $\mathbf{l}_d$, carries the true signal, and the remaining fifteen are assigned at random (without replacement) to fifteen concepts that influence the image.
Every variable in the graph plays one of the following six causal roles:

\begin{itemize}
\itemsep0.2em
\item \emph{True predictive signal} (one variable): This signal is carried by the image attribute $\mathbf{l}_d \in \mathbf{L}$, which drives a path into both the image and the outcome, $\mathbf{X} \leftarrow \mathbf{l}_d \rightarrow \mathbf{y}$. Since $\mathbf{l}_d$ is unobservable, every C-XAI method must use the outcome $\mathbf{y}$ as the observable proxy to estimate the effect size of this pathway.
\item \emph{Confounder} ($2$ concepts): $\mathbf{c}_1$ and $\mathbf{c}_2$ each drive a path into both the image and the outcome, $\mathbf{y} \leftarrow \mathbf{c}_i \rightarrow \mathbf{l}_a \rightarrow \mathbf{X}$ for $i \in \{1,2\}$, where $\mathbf{l}_a$ is the attribute assigned at random to that concept.
\item \emph{Encoded in image} ($13$ concepts; e.g., $\mathbf{c}_j$, $\mathbf{c}_k$, $\mathbf{c}_l$ in \autoref{fig:res-toybrains}b): These concepts have a path into the image and none into the outcome, $\mathbf{c}_i \rightarrow \mathbf{l}_a \rightarrow \mathbf{X}$. Thus, they are genuinely visible in the image but irrelevant to the prediction.
\item \emph{Encoded in outcome} ($3$ concepts): These concepts have a path into $\mathbf{y}$ and none into $\mathbf{X}$, that is, $\mathbf{c}_i \rightarrow \mathbf{y}$ for $i \in \{3,4,5\}$.
\item \emph{Correlated concept} ($60$ concepts; $\mathbf{c}_s$, $\mathbf{c}_t$ in \autoref{fig:res-toybrains}b): These concepts have no edge of their own into $\mathbf{X}$ or $\mathbf{y}$. They are instead resampled to correlate with one of the $15$ image-influencing concepts listed above (the $13$ image-encoded concepts and the two confounders), as $\mathbf{c}_i \leftarrow \mathbf{c}_{s} \rightarrow \mathbf{l}_a \rightarrow \mathbf{X}$. Each of these concepts is assigned one of three correlation strengths at random: $r \in \{0.2, 0.4, 0.6\}$. This results in twenty concepts at each strength, enabling us to study the effect of $r$ on each method (\autoref{fig:res-toybrains}d).
\item \emph{Independent concept} ($22$ concepts; $\mathbf{c}_u$, $\mathbf{c}_v$): The remaining concepts have no edges and are independent of every other variable in the graph. They serve as the $r = 0$ reference in \autoref{fig:res-toybrains}d.
\end{itemize}

Following from this causal graph, the sampling of the binary outcome is given by,
\begin{equation}\label{eq:toybrains_y}
P(\mathbf{y}{=}1) \;=\; \sigma\!\Big(
\beta_{\mathbf{l}_d}\mathbf{l}_d
+ \beta_{\mathbf{c}_1}\mathbf{c}_1 + \beta_{\mathbf{c}_2}\mathbf{c}_2
+ \textstyle\sum_{i=3}^{5}\beta_{\mathbf{c}_i}\mathbf{c}_i \Big)
\end{equation}
where $\sigma$ is the logistic function.

We alter the data-generating process and generate six different causal scenarios by varying $\beta_{\mathbf{l}_d}$, $\beta_{\mathbf{c}_1}$, and $\beta_{\mathbf{c}_2}$, while keeping all other edges in the graph fixed. These settings range from a purely confounder-driven regime ($\beta_{\mathbf{l}_d}=0$) to predominantly signal-driven regimes in which one of the two confounders is switched off, as listed in Extended Data \autoref{tab:ED:toybrains_methods-settings}.

\textbf{Sample sizes.}
For each of the six settings we generate $5{,}000$ training and $5{,}000$ test samples.
DNN models are trained on the training split, while the test split is used as the auditing dataset to generate C-XAI explanations and compute the ground truth.
By default, all methods are applied to a common auditing dataset of $n=500$ images, drawn at random from the test split independently for each model, whereas the ground truth is always computed on the full test split.
Each method is supplied with the same $q=30$ concepts for each model checkpoint, drawn at random from the pool of $100$ concepts, stratified by causal role, ensuring that all causal roles are represented.

\textbf{Ground truth.} C-XAI methods score each concept separately at every layer, so our ground truth must be layer-wise too.
We obtain it by holding the trained weights fixed and intervening on the causal graph in the test distribution (full test split, $n = 5{,}000$).
We define the ground-truth importance of a generative variable $\mathbf{z}$ at layer $l$ as the share of representation variance accounted for by intervening on $\mathbf{z}$:
\begin{equation}\label{eq:gtvar}
\mathrm{Imp}_{\mathrm{true}}(\mathbf{z}) \;=\;
\frac{\mathrm{Var}\!\big(\mathbb{E}\big[\tilde{\mathbf{H}}_l \mid \mathrm{do}(\mathbf{z})\big]\big)}
     {\mathrm{Var}\big(\tilde{\mathbf{H}}_l\big)}
\end{equation}
where $\tilde{\mathbf{H}}_l$ denotes the activation at layer $l$, standardised dimension-wise, and $\mathrm{Var}(\tilde{\mathbf{H}}_l) = \sum_{j} \mathrm{Var}(\tilde{\mathbf{h}}_{l,j})$ denotes the total variance of the standardised representation. 
Here, $\mathrm{do}(\mathbf{z})$ denotes a causal intervention operator \cite{pearl16_causalbook}, so the numerator is the variance in $\tilde{\mathbf{H}}_l$ accounted for by varying $\mathbf{z}$ across its states, averaged over the remaining variables in the image-generating graph.

Only the variables that have a causal path into $\mathbf{X}$ can influence the variance of the model's representations $\mathbf{H}_l$, since intervening on a variable without a directed path into $\mathbf{X}$ leaves the distribution of $\mathbf{X}$, and hence of $\mathbf{H}_l$, unchanged. Of the $100$ concepts, $85$ have no causal path into $\mathbf{X}$. This includes the $60$ correlated concepts, $22$ independent concepts, and $3$ encoded-in-outcome concepts. Therefore, we simply set their ground-truth importance to zero analytically.  
The remaining $15$ concepts and $\mathbf{l}_d$ have a causal path into $\mathbf{X}$. Since they do not have a causal parent, no back-door path runs from any of them into $\mathbf{H}_l$, and the back-door criterion is satisfied by the empty adjustment set~\cite{pearl16_causalbook}.
Therefore, $\mathbb{E}[\tilde{\mathbf{H}}_l \mid \mathrm{do}(\mathbf{z})] = \mathbb{E}[\tilde{\mathbf{H}}_l \mid \mathbf{z}]$, and we can estimate $\mathrm{Imp}_{\mathrm{true}}(\mathbf{z})$ directly from the observational test split. 
Since the $16$ variables are independently sampled, their first-order contributions in \autoref{eq:gtvar} do not overlap, whereas any variance due to interactions between them is not credited to any single variable. 


\textbf{Error from the ground truth.} 
We place all the methods and the ground truth on a common simplex by normalising their scores over the variable set $\mathbf{C}'$ (denoted as $\bar{\mathrm{Imp}}$; \autoref{eq:relimp}).  This makes a distance between them well defined. We then compute the error as the total variation distance ($\mathrm{TVD}$):
\begin{equation}\label{eq:tvd}
\mathrm{TVD} \;=\; \tfrac{1}{2}\sum_{\mathbf{c}_k \in \mathbf{C}'}
\big|\,\bar{\mathrm{Imp}}_{\mathrm{true}}(\mathbf{c}_k) - \bar{\mathrm{Imp}}(\mathbf{c}_k)\,\big|
\end{equation}
where $\bar{\mathrm{Imp}}$ is the normalised score of the evaluated method, computed separately for each method, model and layer.
Since both score vectors sum to one, $\mathrm{TVD}$ is bounded in $[0,1]$ where $0$ denotes exact recovery and $1$ denotes complete disagreement, irrespective of the value of $q$. Thus, the error metric is directly comparable across methods, causal scenarios and layers of differing dimensionality.
One caveat applies to the comparison with CAV and probe baselines. Unlike the ground truth, ICON, or the other baseline methods, their scores are not shares of representation variance, so their TVD also partly reflects the difference in scales.

\textbf{DNN model.} We train a DNN with three convolutional blocks ($32 \to 64 \to 128$ channels, each comprising a $3\times3$ convolution, ReLU activation, batch normalisation and $2\times2$ max pooling), followed by three fully connected layers (dropout $0.1$ between them), referred to as $\mathbf{H}_{L-2}$, $\mathbf{H}_{L-1}$ and $\mathbf{H}_{L}$ (\autoref{fig:res-toybrains}c), where $\mathbf{H}_{L}$ is the final scalar decision logit ($p_L = 1$). We apply all C-XAI methods to these last three layers of the model.
Training uses Adam with initial learning rate $10^{-3}$, batch size $128$, and up to $100$ epochs with early stopping on the validation loss (patience $15$). 
To check that our conclusions are not specific to one architecture or one random initialisation, we train five architectural variants that differ in the widths of the two hidden interpreted layers $(\mathbf{H}_{L-2}, \mathbf{H}_{L-1})$ as $(512, 128)$, $(1024, 256)$, $(2048, 512)$, $(4096, 1024)$ and $(8192, 2048)$, and repeat each variant with three random seeds.
This gives $15$ models per causal scenario and $90$ models across the six scenarios.
The $90$ models reach a test balanced accuracy of $75.8 \pm 2.0\%$ (range $70.7$--$79.1\%$), with means of $75.5$--$76.0\%$ across the five architectural variants.

\textbf{Sensitivity analyses.} We perform three sweeps in which we vary one of the settings fixed above around its default: the auditing dataset size $n$, the concept-set size $q$, and the width of the interpreted layers.
First, we sweep $q \in \{5, 10, 15, 20, 30, 40, 50, 60, 70, 80, 90, 100\}$ at fixed $n = 500$ (Extended Data \autoref{fig:ED:toybrains-pqsweep}a).
Next, we sweep $n \in \{25, 50, 100, 250, 500, 1{,}000, 2{,}500, 5{,}000\}$ at fixed $q = 30$, with the auditing dataset for each $n$ drawn independently (Extended Data \autoref{fig:ED:toybrains-pqsweep}b).
Finally, we compare the five width variants of $(\mathbf{H}_{L-2}, \mathbf{H}_{L-1})$ at fixed $n = 500$ and $q = 30$ (Extended Data \autoref{fig:ED:toybrains-pqsweep}c).  


\subsection{Experiment 2: Skin cancer detection}\label{sec:methods:exp2}

\textbf{Dataset.}
The ISIC 2019 Challenge training set consists of $25{,}331$ dermoscopic images of skin conditions (see examples in \autoref{fig:isic_results}a), classified into nine diagnostic classes, such as the malignant \texttt{melanoma} class (\texttt{MEL}) and the benign \texttt{melanocytic nevus} class (\texttt{NV}) (Extended Data \autoref{fig:ED:isic_data_dist}a, top).
As concepts we use $q{=}9$ artifacts: seven natural artifacts annotated by Pahde~\cite{pahde25_thesis}, such as \texttt{band\_aid} ($n{=}184$) and \texttt{ruler} ($n{=}260$), and two synthetic artifacts, \texttt{microscope} and \texttt{timestamp}, which we insert into the images (Extended Data \autoref{fig:ED:isic_data_dist}a, bottom). The natural artifacts are spuriously correlated with different outcome classes, e.g., \texttt{red\_color} is prevalent in the benign \texttt{NV} class ($r = 0.22$), \texttt{ruler} in the \texttt{MEL} class ($r = 0.18$), and \texttt{reflection} in the benign keratosis class (\texttt{BKL}, $r = 0.16$) (Extended Data \autoref{fig:ED:isic_data_dist}b). 

\textbf{Sample sizes.}
Of the $25{,}331$ ISIC images, we exclude the $n{=}795$ images that contain a natural \texttt{microscope} lens artifact.
We split the remaining $n{=}24{,}536$ images randomly into training ($80\%$, $n{=}19{,}622$), validation ($10\%$, $n{=}2{,}455$) and test ($10\%$, $n{=}2{,}459$) sets, stratified by the outcome with a fixed seed. 

\textbf{DNN Model.} 
We fine-tune an ImageNet-pretrained VGG-16~\cite{simonyan14_vgg16} on the training split with SGD (learning rate $0.001$, reduced by a factor of $0.1$ at epochs 100 and 150 of 200; batch size $128$) and a class-weighted cross-entropy loss. Training images are augmented with random flips, blur, rotation and colour jitter. 
We use the validation split to monitor convergence and select the best checkpoint for explanations. 
C-XAI explanations are computed on the test split at six layers, from the convolutional layer $\mathbf{H}_{L-17}$ to the prediction logits $\mathbf{H}_L$ (\autoref{fig:isic_qualitative}), and the quantitative tests in \autoref{fig:isic_results} use the last two, $\mathbf{H}_{L-1}$ and $\mathbf{H}_L$. 

\textbf{Synthetic artifacts.}
Similarly to Pahde~\cite{pahde25_thesis}, we induce a spurious correlation between an artifact and an outcome class by inserting a synthetic artifact into a share of that class's images (rightmost examples in \autoref{fig:isic_results}a).
We denote such an adulterated dataset variant as, e.g. $\mathcal{D}_{\text{micro-}k}$, where $k$ is the percentage of the \texttt{MEL} samples in which the \texttt{microscope} artifact was inserted. In contrast,  $\mathcal{D}_{\text{micro-00}}$ is the original ISIC dataset above that excludes all natural \texttt{microscope} artifact samples too.
DNN models trained on the adulterated datasets are denoted as $\mathcal{M}_{\text{micro-}k}$. 
These models learn to rely on the synthetic concept as a shortcut to different extents depending on the value of $k$. 
We quantify how much $\mathcal{M}_{\text{micro-}k}$ relies on the artifact as $\Delta \text{acc} = \text{acc}(\mathcal{M}_{\text{micro-}k}, \mathcal{D}_{\text{micro-}k}) - \text{acc}(\mathcal{M}_{\text{micro-}k}, \mathcal{D}_{\text{micro-00}})$, i.e., the difference in the balanced accuracy when the model is evaluated on its own (adulterated) test data versus on the test set of the clean (unadulterated) data. 

We create nine dataset variants, $\mathcal{D}_{\text{micro-10}}$ to $\mathcal{D}_{\text{micro-90}}$, in steps of $10\%$, and train one model on each. 
$\Delta\text{acc}$ increases monotonically from $+0.2$ ($\mathcal{M}_{\text{micro-10}}$) to $+7.0$ percentage points ($\mathcal{M}_{\text{micro-90}}$) (\autoref{fig:isic_results}b, bottom). 

We use these models to test the specificity and responsiveness of ICON, linear probes, CAV-classic and CAV-signal in three tests, one per column of \autoref{fig:isic_results}.
The first tests whether a method tracks increasing degrees of shortcut learning.
The second and third test whether it produces false positives when the concept--outcome or the concept--concept collinearity changes in the auditing data.
Such distribution shifts are common in practice, since spurious correlations in the training data may be absent at a clinical site where the model is deployed~\cite{geirhos20_shortcutlearninginDL}.

\textbf{Shortcut strength (\autoref{fig:isic_results}b).}
We explain \texttt{microscope} in all ten models, $\mathcal{M}_{\text{micro-00}}$ to $\mathcal{M}_{\text{micro-90}}$, on the same auditing dataset, the test split of $\mathcal{D}_{\text{micro-10}}$, so that the auditing data are identical across models and the \texttt{MEL}--\texttt{microscope} correlation is low ($r = 0.30$).
A reliable C-XAI method should assign \texttt{microscope} an importance $\bar{\mathrm{Imp}}(\texttt{microscope})$ that increases with $\Delta\text{acc}$.

\textbf{Concept--outcome collinearity (\autoref{fig:isic_results}c).}
We explain \texttt{microscope} in the clean model $\mathcal{M}_{\text{micro-00}}$ on the test splits of $\mathcal{D}_{\text{micro-10}}$ to $\mathcal{D}_{\text{micro-90}}$, in which the \texttt{MEL}--\texttt{microscope} correlation rises from $0.30$ to $0.94$ (\autoref{fig:isic_results}c, bottom).
Since $\mathcal{M}_{\text{micro-00}}$ was trained without \texttt{microscope} artifacts, it cannot have learned to use the artifact, so we treat the ground-truth importance of \texttt{microscope} in its last two layers as zero, whatever the artifact's prevalence in the auditing data.
This criterion does not apply to the early convolutional layers, where the inserted artifact is present in the input and is therefore expected to be encoded (\autoref{fig:isic_qualitative}).

\textbf{Concept--concept collinearity (\autoref{fig:isic_results}d).}
We insert a second synthetic artifact, \texttt{timestamp}, into $k\%$ of the \texttt{microscope}-carrying images of $\mathcal{D}_{\text{micro-40}}$, giving the variants $\mathcal{D}_{\text{micro-40+time-}k}$, whose test splits serve as auditing datasets in which the \texttt{microscope}--\texttt{timestamp} correlation rises from $0.32$ to $0.95$ (\autoref{fig:isic_results}d, bottom).
We explain \texttt{timestamp} in $\mathcal{M}_{\text{micro-40}}$, which relies on \texttt{microscope} to some extent ($\Delta\text{acc} \approx 1.3$ percentage points).
Because \texttt{timestamp} appears in no training set, we again treat its ground-truth importance in the last two layers as zero.

\subsection{Experiment 3: Neuroimaging applications}\label{sec:methods:exp3}
 Experiment 3 comprises two prediction tasks on UK Biobank brain MRI: binge-drinking classification (Experiment 3a) and brain-age regression (Experiment 3b).
 
\textbf{Data and preprocessing.}
We use T1-weighted structural MRI from the UK Biobank~\cite{sudlow15_ukbb}, preprocessed with the standard UK Biobank pipeline (skull-stripping and non-linear registration to MNI-152 space)~\cite{alfaro18_ukbbpipeline} and then cropped and resampled to $96\times114\times96$ voxels.
Both tasks use five random 80\%/20\% train/validation splits of the training cohort, stratified by the outcome.
 
\textbf{Concept set.}
As listed in Extended Data \autoref{tab:ED:neuro_concepts}, both tasks are interpreted with 30 concept variables spanning demographics, cognitive tests, self-reported mental health, physical health, socioeconomic status, and MRI acquisition settings. After appending the respective outcome variable to the concept set, $| \mathbf{C}'|$ becomes 31 variables.  
We standardise continuous variables and one-hot encode categorical variables.
Several concepts are strongly intercorrelated (Extended Data \autoref{fig:ED:neuro_corr}), as is common in neuroimaging cohorts~\cite{alfaro21_ukbbconfound}.
 
\textbf{C-XAI interpretation.} For each model, we interpret six layers, from an early convolutional layer to the prediction $\mathbf{H}_L$: $\mathbf{H}_{L-20}$, $\mathbf{H}_{L-16}$, $\mathbf{H}_{L-11}$, $\mathbf{H}_{L-6}$, $\mathbf{H}_{L-1}$ and $\mathbf{H}_L$ for the ResNet-18, and $\mathbf{H}_{L-7}$, $\mathbf{H}_{L-5}$, $\mathbf{H}_{L-3}$, $\mathbf{H}_{L-2}$, $\mathbf{H}_{L-1}$ and $\mathbf{H}_L$ for the SFCN (\autoref{fig:neuro}). On each layer, we compute ICON importance and the normalised linear-probe importance (\autoref{eq:relimp}) on the held-out auditing dataset.
We do not compare against CAV-based methods here, because TCAV~\cite{kim18_TCAV} and CAV-signal~\cite{pahde24_signalCAV} are defined for binary concepts and RCAV~\cite{graziani2018_RCAV} for continuous ones, so they cannot score our mixed concept set on one comparable scale. In addition, in the skin-cancer experiment we already reproduced the previously reported instability of CAV-based metrics~\cite{nicolson25_CAVcritique, ramaswamy23_CAVtestdata} (\autoref{fig:isic_results}c,d).
 
\textbf{Binge-drinking classification (Experiment 3a).}
The outcome variable is derived from the UK Biobank's self-reported binge-drinking frequency field (Field ID 20416). To binarise it, we use a threshold analogous to that of the reference study~\cite{rane2022structural}: participants who report binge drinking `weekly' or `daily or almost daily' form the positive class and participants who report `never' form the controls. 
From this pool, we draw two subsamples of $n \approx 6000$ each, with matched age distributions, that differ in their sex--outcome correlation.
The first retains the cohort's natural correlation (Pearson $r(\texttt{sex},\text{outcome})\approx0.23$) and the second is balanced by sex, so that $r(\texttt{sex},\text{outcome})\approx0.00$.
On each subsample, we train a 3D ResNet-18~\cite{hara18_3dresnets}, pretrained on video data, with identical hyperparameters: Adam (learning rate $1.1\times10^{-5}$, weight decay $2.7\times10^{-6}$, both tuned with Optuna~\cite{akiba2019optuna}), batch size 32, and up to 120 epochs with early stopping (patience 15).
The natural-correlation model reaches an accuracy of $60.0\pm0.5\%$  and the sex-balanced model only $53.3\pm0.5\%$ on their respective $n \approx 1{,}200$ hold-out participants (chance $50\%$).
Because the two subsamples differ mainly in their sex--outcome correlations, this gap is an estimate of how much of the accuracy depended on that correlation.
\autoref{fig:neuro}a shows the natural-correlation model.
To check for architecture-specific effects, we also train a Simple Fully Convolutional Network (SFCN)~\cite{peng19_brainage} from scratch on the natural-correlation subsample, under the same training and tuning protocol, and compare the ICON explanations of the two architectures (Extended Data \autoref{fig:neuro_diagnostics}b).
 
\textbf{Brain-age prediction (Experiment 3b).}
Here the target is chronological age (regression).
For training, we sample $n=10{,}000$ `healthy' participants (mean age $54.7\pm7.1$ years), excluding anyone with a prior neurological or psychiatric diagnosis (ICD-10 chapters F and G), as is standard in the brain-age-gap paradigm~\cite{cole19_brainagereview}. The $16{,}661$ participants not used for training, including those with such diagnoses, form the hold-out pool. From it, we draw a fixed test set of $n=5{,}000$ participants (mean age $54.4\pm7.5$ years), which serves as the auditing dataset and as the reference age distribution for Validation~1, whose OOD and matched-control sets are drawn from the same pool. 
We use the SFCN, which won the 2019 Predictive Analytics Competition (PAC) for brain-age prediction~\cite{peng19_brainage}, and train it with Adam (learning rate $1.5\times10^{-3}$, weight decay $2.7\times10^{-8}$; both selected with Optuna~\cite{akiba2019optuna}), batch size 32, and a \texttt{ReduceLROnPlateau} schedule for up to 150 epochs with early stopping (patience 15) on the validation loss; the age target is standardised before fitting.
On the fixed test set, the model reaches a mean absolute error (MAE) of $2.89\pm0.02$ years and $R^2=0.750\pm0.004$.
To show how concept importance evolves during training, we also apply ICON to three checkpoints of fold~2, at initialisation, at epoch $13$ and at convergence (epoch $97$) (Extended Data \autoref{fig:neuro_diagnostics}a).
 
\textbf{Independent validation of the brain-age explanations.}
We use two complementary validation tests to determine whether the brain-age model relies on the concepts flagged by ICON or the linear probe.
First, resampling-based out-of-distribution (OOD) tests assess whether the model relies on a concept as a proxy for the outcome.
Second, a conditional independence test (CIT) detects any dependence of the prediction on a concept beyond the outcome and the other selected concepts.
For the testing, we take the top five ranked concepts of ICON and probes (other than \texttt{age}) at the prediction layer $\mathbf{H}_L$, after averaging the importances across folds. The union of the two top-five lists gives eight concepts listed in \autoref{tab:neuro_ood}.
The probe ranks \texttt{employment\_status} as the most important concept after age and ICON ranks \texttt{acquisition\_date}, whereas \texttt{snap\_reaction\_time} and \texttt{systolic\_blood\_pressure} appear in the top five of both. 
 
\emph{Validation 1: matched-resampling OOD test.}
For each selected concept, we draw two disjoint, equally sized sets from the hold-out pool: an OOD set in which the concept's correlation with age is removed, and a matched control set in which it keeps its natural value (Extended Data \autoref{fig:neuro_ood_corr}).
We ensure that both sets match in their age distribution by binning them into 2-year age bins and drawing the same number of participants per bin for the two sets, in proportion to the original test set's bin frequencies.
Within each bin, the OOD set additionally balances participants across the concept's categories (or global quantile bins, for continuous concepts), which drives its correlation with age to near zero.
The error gap $\Delta\text{MAE}=\text{MAE}_{\text{OOD}}-\text{MAE}_{\text{control}}$ therefore indicates how much the model's accuracy depends on that correlation.

We expect $\Delta\text{MAE}>0$ if the model \textit{relies} on the concept as a proxy for age, whereas a false-positive concept should show $\Delta\text{MAE}\approx0$.
We set the following sampling thresholds to ensure that the two sets are reasonably matched: (a) the residual concept-age correlation stays below $0.03$, (b) age mean and variance match the reference test data to within $0.05$ years and $1\%$, and (c) the final sample size is $n\ge200$. 
Two of the eight concepts do not pass these thresholds. For \texttt{acquisition\_date} and \texttt{employment\_status}, some age bins contain no participants in at least one of the concept's categories (for example, no retired participants in the youngest bins), so no candidate set could be constructed. 
We test each $\Delta\text{MAE}$ per fold with a two-sided bootstrap. 
Since the folds reuse the same participants and are therefore not independent, we call a concept significant only if its largest $P$ value across the five folds falls below $\alpha=0.05/6=0.0083$, Bonferroni-corrected over the six tested concepts (Methods~\ref{sec:methods:stats}). 

\emph{Validation 2: CIT.}
As a second check, we test the null hypothesis $H_0 : \mathbf{H}_L \perp \mathbf{c}_k \mid Z_k$, where $Z_k=\{\texttt{age}\}\cup\{\mathbf{c}_i : i\neq k\}$. 
That is, we test if the prediction $\mathbf{H}_L$ is conditionally independent of a concept $\mathbf{c}_k$ given the outcome and the remaining seven selected concepts.

As CIT we apply the generalised covariance measure~\cite{shah20_GCM}, which regresses $\hat{\mathbf{y}}$ on $Z_k$ and $\mathbf{c}_k$ on $Z_k$ separately and then asks whether the two sets of residuals are uncorrelated.
If the prediction still tracks the concept after both have been stripped of everything $Z_k$ explains, the concept carries information about the prediction that the outcome and the other concepts do not.
We use the \texttt{comets} R package~\cite{kook24_comets} implementation where regressions are random forests (500 trees), which accommodate the mixed continuous, ordinal and binary variables in $Z_k$.

The test uses the $n=3{,}738$ test-set participants with annotations for all eight concepts and is run once per fold with that fold's predictions.
A concept counts as significant only if its largest $P$ value across the five folds falls below $\alpha=0.05/8=0.0063$ (Methods~\ref{sec:methods:stats}).

%% file: sections/supplementary.tex
\section*{Supplementary Note 1: Literature review of concept-based explainability methods}
\label{A:cxai_history}

Concept-based explainability (C-XAI) methods quantify how important a human-understandable \emph{concept} such as patient sex, a disease indicator, or an imaging artifact, is to a deep neural network (DNN), by analysing the network's internal representations.
They have gained popularity due to their flexibility as they can interpret any trained DNN model in terms of concepts a domain expert chooses, when annotations are available. Furthermore, this can inform post-hoc model auditing and model steering features \cite{poeta23_ConceptXAIsurvey, gupta24_ConceptXAIsurvey, erogullari25_disentangleCAVs}.
The two dominant approaches are linear probes and Concept Activation Vectors (CAVs).
We review each below and show that both inherit the same \emph{univariate decoding} limitation that motivates ICON.

\paragraph{Linear probes.}
A linear probe is a linear model trained on a DNN layer's representation to predict a concept. The probe's accuracy is then used as a measure of how strongly that concept is encoded in that representation \cite{alain16_probing, belinkov22_probingsurvey}.
The central problem is that \emph{decodability does not imply use}: a concept can be linearly decodable from a layer without the model ever relying on it to make predictions.
Ravichander et al.~\cite{ravichander20_probingcorrelation} show that probes recover concepts with high accuracy even when those concepts are irrelevant to the task, so probing accuracy alone cannot establish importance.
Elazar et al.~\cite{elazar21_probingamnesic} make the same point causally: removing a property from the representation through an ``amnesic'' intervention often leaves the model's predictions unchanged, confirming that decodability and behavioural relevance are distinct.
Using probe accuracy as an importance score therefore produces false positives for concepts that are merely correlated with the representation or with the outcome.

\paragraph{Concept Activation Vectors.}
CAV-based methods were introduced to close this specific gap between decodability and use, by measuring not only if a concept is decodable, but also how sensitive the model's output is to it.
A CAV is the direction in activation space separating concept from non-concept examples; TCAV \cite{kim18_TCAV} then scores importance using the directional derivative of the output along this direction, and Regression Concept Vectors (RCAV) \cite{graziani2018_RCAV} extend the idea to continuous concepts.
Because the score relies on the model's gradient, a concept the model does not use should receive a low importance score.
In practice, however, CAV-based scores inherit two problems, the first of which is instability.
The concept importance scores can shift substantially under small changes to the concept's training examples and to the choice of auditing dataset \cite{ramaswamy23_CAVtestdata}, and under adversarial perturbations of the inputs \cite{brown21_CAVadversarial}. When the same concept is measured at different adjacent layers of the same network, they often generate inconsistent importance scores \cite{nicolson25_CAVcritique}.
The second problem, which CAVs share with probes, is that a concept's score absorbs the contribution of the concepts it is correlated with.

\paragraph{Concept entanglement.}
A failure common to both probes and CAVs is \emph{concept entanglement}: because a layer's representation mixes many correlated concepts, an estimate for one concept can absorb the contribution of others that co-occur with it \cite{nicolson25_CAVcritique, raman24_CAVcorrelation}.
Erogullari et al.~\cite{erogullari25_disentangleCAVs} demonstrate this directly on the CelebA dataset, where ``beard'' and ``necktie'' co-occur almost exclusively in images of men: CAVs trained for the two concepts point in nearly the same (non-orthogonal) direction, so steering a generative model along the ``necktie'' direction also introduces correlated facial-hair concepts such as a moustache.
The two concepts cannot be told apart from the representation alone, and any univariate score will attribute their shared variance to both.
This is precisely why a concept the model never used can still be flagged as important whenever it is correlated with a concept (or with the outcome) that the model did use.

Several refinements have been proposed: signal-based CAVs that recover more faithful concept directions by moving away from decodability~\cite{pahde24_signalCAV}, orthogonalisation that separates correlated concept directions after training~\cite{erogullari25_disentangleCAVs}, and usage recommendations and correlation-aware diagnostics for interpreting CAVs~\cite{nicolson25_CAVcritique, raman24_CAVcorrelation}.
Each addresses an individual symptom such as direction noise, layer sensitivity, or pairwise entanglement, rather than the shared root cause.

That root cause is the univariate decoding framing itself. Since probes and CAVs analyse one concept at a time, they cannot separate a concept's unique contribution from the contributions of all other correlated concepts and the outcome.
ICON addresses this directly by reframing C-XAI as a multivariate variance-decomposition problem, in which all concepts are evaluated jointly and shared variance is allocated explicitly rather than counted twice (see main text).


%% file: sections/extended_data.tex

\section{Extended Data}
\setcounter{figure}{0}
\setcounter{table}{0}
\renewcommand{\figurename}{Extended Data Fig.}
\renewcommand{\tablename}{Extended Data Table}

\begin{table}[ht]
    \centering
    \resizebox{\textwidth}{!}{%
        \input{figures/toybrains_methods-settings}
    }\caption{\textbf{ToyBrains simulation (Experiment 1). Causal scenarios, model accuracy and layer-wise ground-truth importance.}
We tune the three causal edges $\beta_{\mathbf{l}_d}$, $\beta_{\mathbf{c}_1}$ and $\beta_{\mathbf{c}_2}$ into $\mathbf{y}$ in the data-generating graph (\autoref{fig:res-toybrains}b; \autoref{eq:toybrains_y}) to emulate six causal scenarios, ranging from purely confounder-driven, through mixtures of the true signal with one or both confounders, to predominantly signal-driven.
Note that the ranking of $\mathbf{c}_1$ and $\mathbf{c}_2$ need not follow their $\beta$ values, because a confounder's empirical importance also depends on its edge into an image attribute, its data type, and on its number of states. Hence, we estimate the ground truth empirically (Methods~\ref{sec:methods:exp1}).
For each scenario we train $15$ models (five architectures $\times$ three seeds) and report their mean balanced accuracy (BAC) on the test split, with the range across models in brackets.
The remaining columns report the ground-truth importance $\bar{\mathrm{Imp}}_{\mathrm{true}}$ (\autoref{eq:gtvar}) of each causal role at the last three layers $\mathbf{H}_{L-2}$, $\mathbf{H}_{L-1}$ and $\mathbf{H}_L$. 
Values are rescaled to percentages of the summed ground-truth importance over all $100$ concepts and the true signal. We show medians $\pm$ s.d. over the $15$ models trained per scenario, where each model is scored against its own ground truth.
Across all scenarios, the share of the $13$ image-encoded concepts falls from $72$--$84\%$ at $\mathbf{H}_{L-2}$ to $1$--$2\%$ at $\mathbf{H}_L$, showing that the networks progressively discard image content that is irrelevant to the prediction.
The roles `encoded in outcome', `correlated concept' and `independent concept' have no causal path into the image and therefore zero ground-truth importance by construction.
}\label{tab:ED:toybrains_methods-settings}
\end{table}

\begin{figure}[!htbp]
    \centering
    \includegraphics[width=\textwidth]{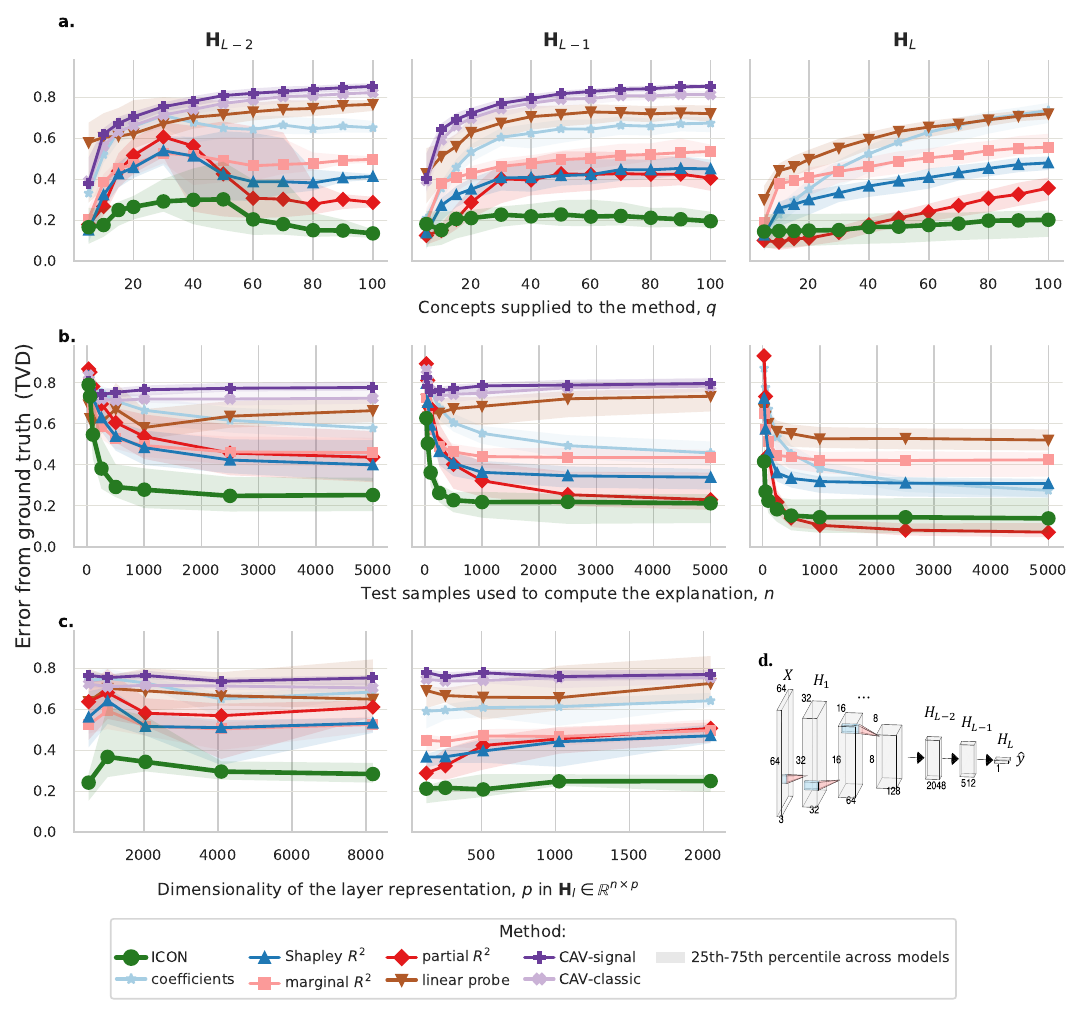}
    \caption{\textbf{ToyBrains simulation (Experiment 1). Sensitivity to concept-set size, auditing sample size and layer width.}
    Error from the ground truth (TVD, \autoref{eq:tvd}) at the last three layers.
    Lines are medians over the $90$ models, with 25th to 75th percentile bands.
    \textbf{a,}~Varying the number of provided concepts over $q \in \{5, 10, 15, 20, 30, 40, 50, 60, 70, 80, 90, 100\}$ at fixed $n = 500$. ICON's error stays at or below $\approx0.3$ at every layer, whereas every baseline's error rises with $q$, since each added concept is one more collinear distractor that the baselines fail to reject. 
    \textbf{b,}~Varying the number of auditing samples over $n \in \{25, 50, 100, 250, 500, 1{,}000, 2{,}500, 5{,}000\}$, each drawn independently, at fixed $q = 30$. ICON's error levels off earliest, at about $n = 250$, whereas partial $R^2$ needs about $n = 5{,}000$ to reach it at $\mathbf{H}_{L-1}$. 
    \textbf{c,}~Varying the width $p_l$ of the interpreted layer, at fixed $n = 500$ and $q = 30$. The five points are the five width variants of $(\mathbf{H}_{L-2}, \mathbf{H}_{L-1})$, from $(512, 128)$ to $(8192, 2048)$. Every method's error varies only modestly with width, and ICON is the most accurate method at every width in both internal layers. 
    The prediction layer $\mathbf{H}_{L}$ is omitted because its width cannot be varied from $p_L = 1$. 
    \textbf{d,}~Architecture of the trained network, showing the variant with widths $(2048, 512)$ for $(\mathbf{H}_{L-2}, \mathbf{H}_{L-1})$.}
    \label{fig:ED:toybrains-pqsweep}
\end{figure}
\begin{figure}[!htbp]
    \centering
    \includegraphics[width=\textwidth]{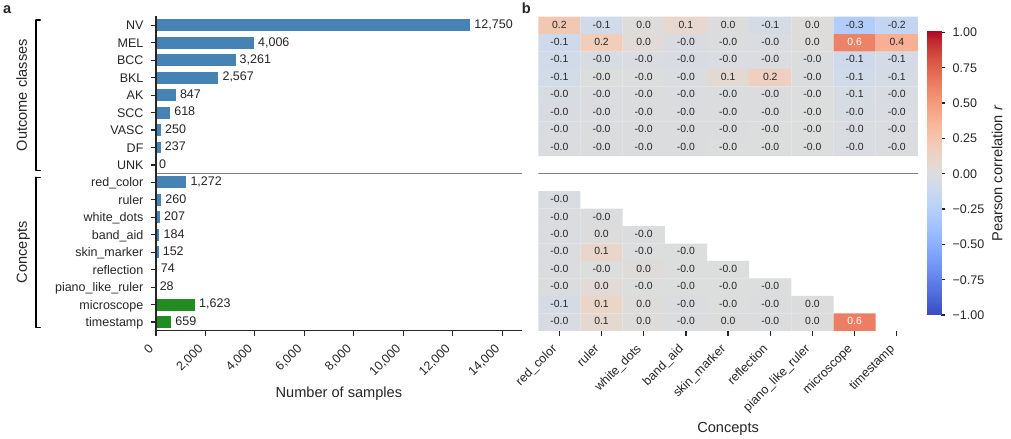}
    \caption{\textbf{Composition of the ISIC 2019 dataset and the correlations among its outcome classes and artifact concepts.}
    \textbf{a,}~Histogram of the nine outcome classes (top) and the nine artifact concepts (bottom).
    The seven natural artifacts (blue bars) are \texttt{band\_aid}, \texttt{ruler}, \texttt{red\_color}, \texttt{white\_dots}, \texttt{skin\_marker}, \texttt{reflection} and \texttt{piano\_like\_ruler}.
    The two synthetic artifacts (green bars), \texttt{microscope} and \texttt{timestamp}, appear only in the adulterated variants; the counts shown are those of $\mathcal{D}_{\text{micro-40}}$.
    \textbf{b,}~Pearson correlations between every pair of the nine outcome classes and the nine artifact concepts, including the correlations of the artifact concepts with one another.
    The natural artifacts are also correlated with the outcome classes, for example \texttt{red\_color} with \texttt{NV} ($r = 0.22$), \texttt{ruler} with \texttt{MEL} ($r = 0.18$) and \texttt{reflection} with \texttt{BKL} ($r = 0.16$).
    This is the background collinearity against which every C-XAI method in \autoref{fig:isic_results} has to isolate the inserted artifact.}
    \label{fig:ED:isic_data_dist}
\end{figure}

\begin{figure}[!htbp]
    \centering
    \includegraphics[width=\textwidth]{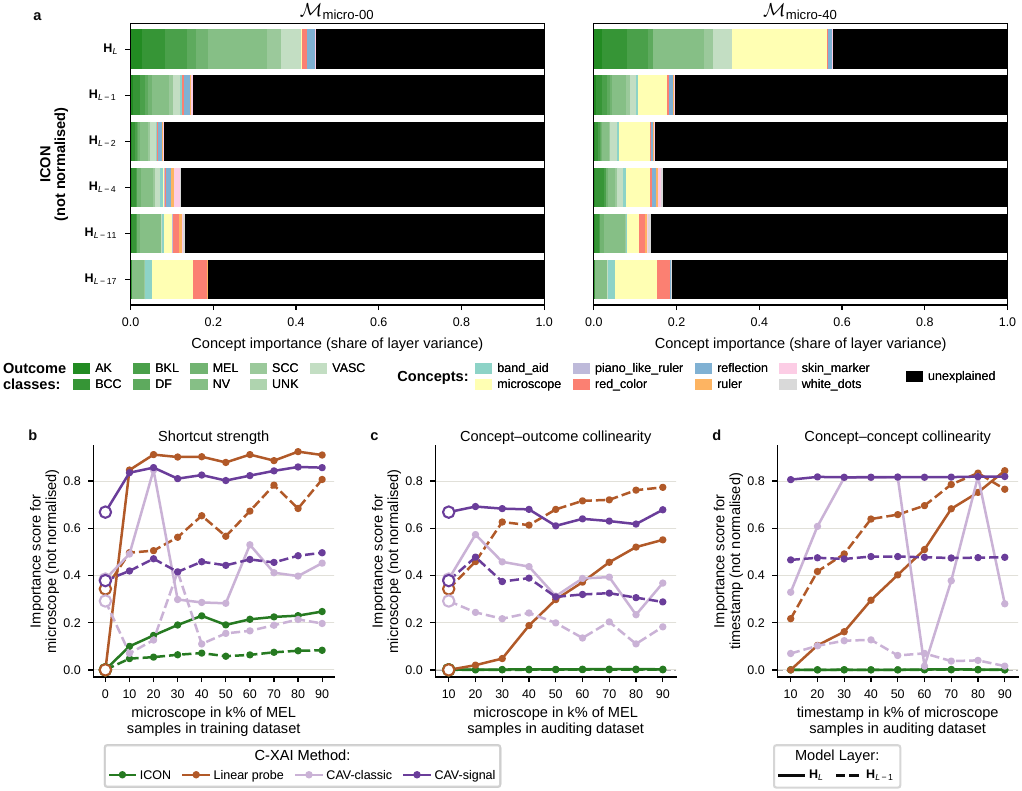}
    \caption{\textbf{Skin-cancer results before normalisation.}
    \textbf{a,}~Layer-wise ICON decomposition of $\mathcal{M}_{\text{micro-00}}$ (left) and $\mathcal{M}_{\text{micro-40}}$ (right), as in \autoref{fig:isic_qualitative} but before normalisation.
    The black segment is the share $\sigma^2_l$ of each layer's variance that none of the nine artifact concepts and nine outcome classes explains (\autoref{eq:icon_step1}); removing it and rescaling the other segments to sum to one gives the ICON row of \autoref{fig:isic_qualitative}.
    The unexplained share is about $0.8$ or more at every hidden layer and falls substantially only at the prediction logits $\mathbf{H}_L$ ($\approx0.55$ for $\mathcal{M}_{\text{micro-00}}$ and $\approx0.42$ for $\mathcal{M}_{\text{micro-40}}$).
    \textbf{b--d,}~The three tests of \autoref{fig:isic_results}b--d with the not normalised scores $\mathrm{Imp}$ in place of $\bar{\mathrm{Imp}}$ (\autoref{eq:relimp}).
    Before normalisation, the linear probe and CAV-signal scores also rise with $k$ in \textbf{b}, mostly between $k = 0$ and $k = 10$, whereas in \textbf{c} and \textbf{d} ICON stays near zero and the three baselines do not.
    Not normalised scores are not comparable in magnitude across methods, because ICON scores are shares of layer variance whereas probe and CAV scores are fit and sensitivity statistics (Methods~\ref{sec:methods:setup}).
    Each point is one model audited on one test split ($n = 2{,}459$ images); open circles mark the same model and auditing dataset in \textbf{b} and \textbf{c}.}
    \label{fig:isic_results_abs}
\end{figure}

 
 
\begin{table}[htbp]
\centering
\footnotesize
\caption{\textbf{Concept set for the neuroimaging experiments.}
The concept variables supplied to ICON and the linear probe, grouped by category, with each variable's type and UK Biobank cohort summary statistics.
Both experiments draw on this shared pool of $31$ variables and supply $30$ of them as concepts.
Each experiment sets aside the variable used to derive its own outcome: brain-age prediction sets aside chronological age, and binge-drinking classification sets aside frequency-of-alcohol-use.
Statistics are computed over the union of the training and hold-out subjects for which concept annotations were supplied in the two experiments ($n=17{,}713$ unique subjects; brain-age cohort $n=15{,}000$, binge-drinking cohort $n=6{,}255$, with $3{,}542$ subjects in both). 
Continuous variables are standardised to zero mean and unit variance; categorical and binary variables are one-hot encoded.}
\label{tab:ED:neuro_concepts}
\begin{tabularx}{\textwidth}{@{} >{\hsize=1.05\hsize\raggedright\arraybackslash}X >{\hsize=0.85\hsize\raggedright\arraybackslash}X p{.12\textwidth} >{\hsize=1.1\hsize\raggedright\arraybackslash}X @{}}
\toprule
\textbf{Concept} & \textbf{Category} & \textbf{Type} & \textbf{Statistics} \\
\midrule
\input{figures/neuro_data-concepts.tex}
\end{tabularx}
\end{table}

\begin{figure}[!htbp]
    \centering
    \includegraphics[width=\textwidth]{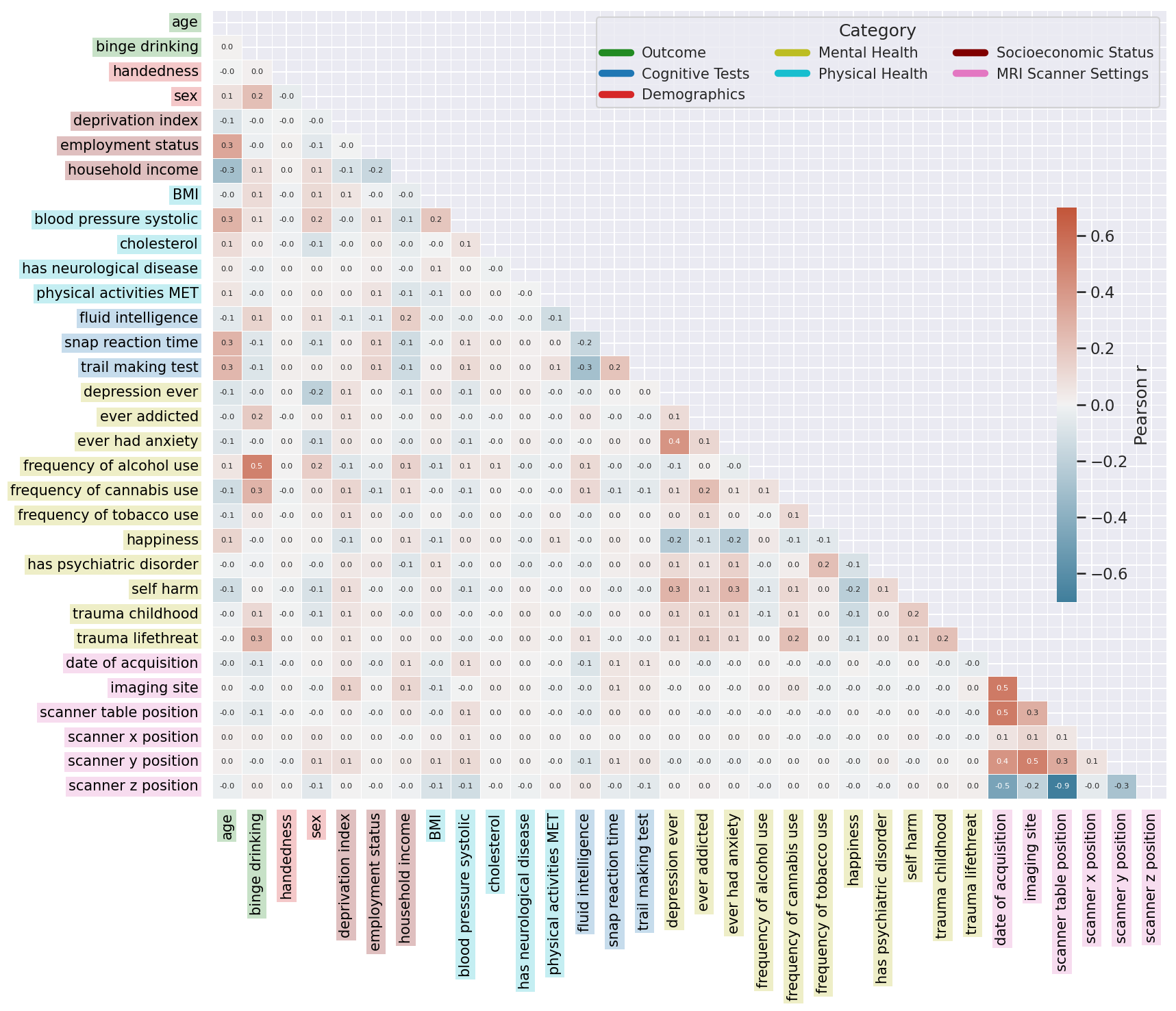}
    \caption{\textbf{Concept and outcome intercorrelations in the UK Biobank cohort.}
    Pearson correlation matrix over all $32$ variables used across the two tasks, the $31$ variables listed in Extended Data \autoref{tab:ED:neuro_concepts} plus the binge-drinking outcome, computed over the $n = 17{,}713$ subjects.
    Each cell is coloured by the sign and magnitude of the pairwise Pearson correlation (diverging scale), and concepts are grouped along both axes into the six categories used throughout: demographics, socioeconomic status, physical health, cognitive tests, self-reported mental health, and MRI scanner-acquisition settings.
    The concepts are strongly intercorrelated: for example, age correlates with several socioeconomic, physical-health and cognitive variables ($|r|>0.2$).
    The scanner-acquisition variables are also highly intercorrelated, with acquisition date and scanner table position reaching $|r| \approx 0.5$ and so are also the self-reported mental-health items.}
    \label{fig:ED:neuro_corr}
\end{figure}

\begin{figure}[!htbp]
    \centering
    \includegraphics[width=\textwidth]{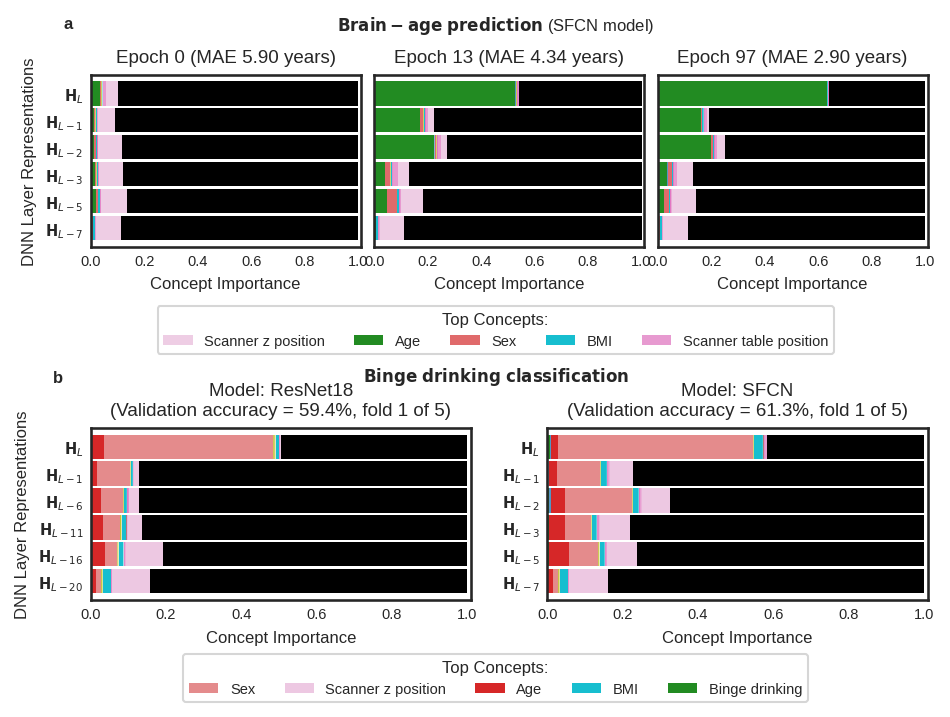}
    \caption{\textbf{ICON tracks how a representation forms over training and compares two architectures on a common scale.}
    In both panels the bars are layer-wise ICON importance, ordered from an input layer at the bottom to the prediction logit $\mathbf{H}_L$ at the top, and the black segment is the share of representation variance that none of the $30$ supplied concepts explains.
    \textbf{a,}~The brain-age SFCN model at three stages of training, all from fold 2 of 5: at initialisation (validation MAE $=5.90$ years), at epoch $13$ (MAE $=4.34$ years) and at convergence at epoch $97$ (MAE $=2.90$ years). Scanner-position concepts dominate the early-layer representation near initialisation, and age comes to dominate the final and penultimate layers as the model converges, so these layers learn to extract the age signal and filter out the scanner information present in the input images.
    \textbf{b,}~The binge-drinking task for a video-pretrained 3D ResNet-18 (left; fold 1 of 5, test accuracy $59.4\%$; $60.0 \pm 0.5\%$ over five folds) and for an SFCN (right; fold 1 of 5, test accuracy $61.3\%$; $61.2 \pm 0.7\%$ over five folds), both evaluated on the $n = 1{,}246$ hold-out participants. Because ICON expresses importance as a proportion of representation variance, the two architectures are directly comparable. \texttt{Sex} explains a comparably large share of the final-layer representation in both ($\approx0.40$ for the ResNet-18 and $\approx0.46$ for the SFCN), so changing the architecture and the pretraining does not substantially reduce the model's reliance on the sex confound at convergence.}
    \label{fig:neuro_diagnostics}
\end{figure}


\begin{figure}[!htbp]
    \centering
    \includegraphics[width=\textwidth]{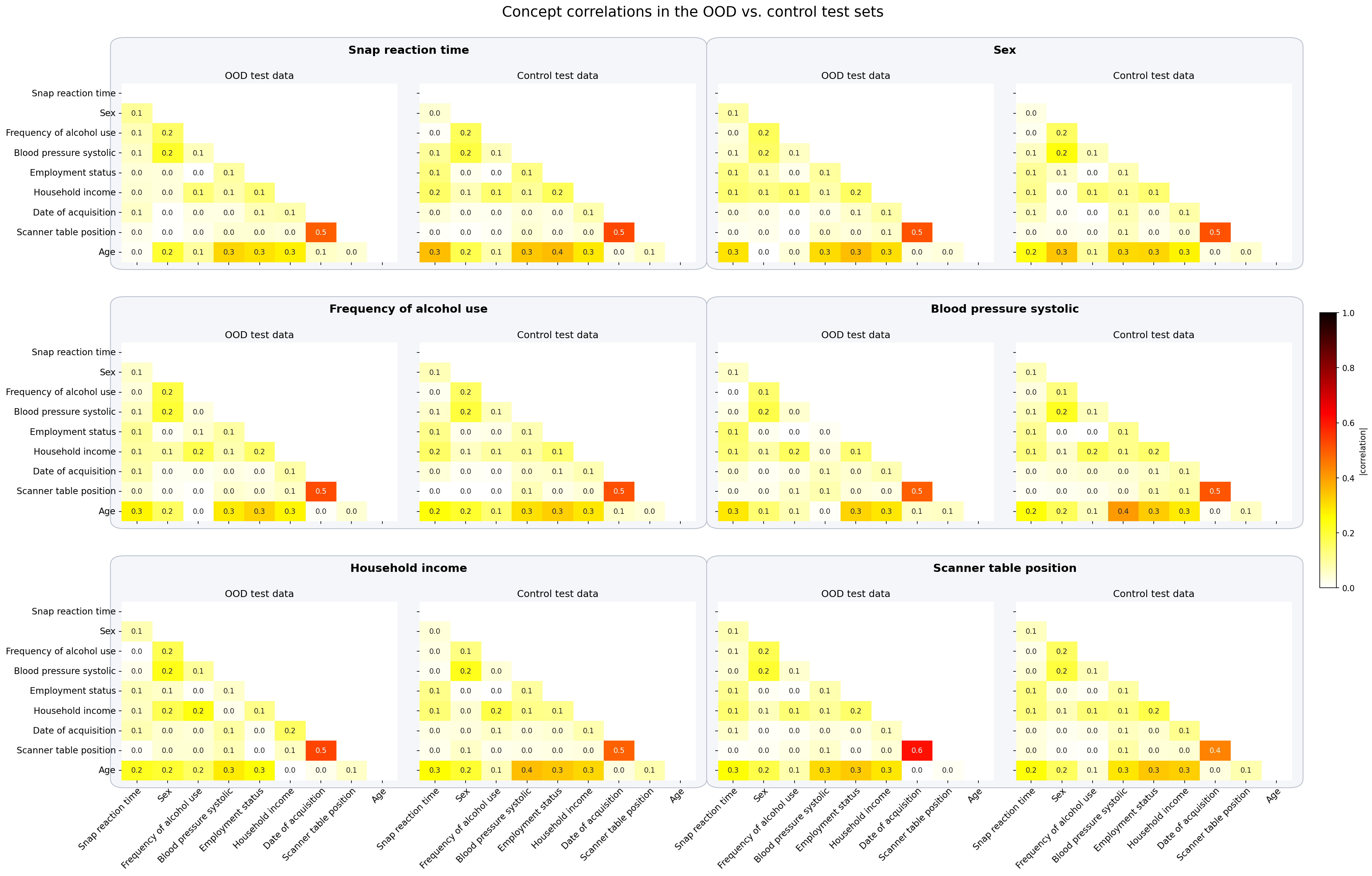}
    \caption{\textbf{Matched resampling removes each concept's correlation with age while leaving the rest of the correlation structure intact.}
    For each of the six concepts that admitted a valid out-of-distribution (OOD) set (\autoref{tab:neuro_ood}), 
    we show the absolute pairwise correlations among all validation concepts and the outcome, age, in the OOD test set (left) versus the matched random-control set (right).
    In every OOD set the target concept's correlation with age (bottom \texttt{age} row) is driven to $\approx0$, whereas the control retains its natural value (for example, $\approx0.4$ for systolic blood pressure and $\approx0.3$ for sex and household income).
    All other correlations, including the strong link between the two scanner variables, acquisition date and scanner table position ($\approx0.5$), are preserved across the two sets.
    Decorrelating the target concept from age is therefore the only intended difference between the OOD and control sets (\autoref{sec:methods:exp3}).}
    \label{fig:neuro_ood_corr}
\end{figure}

%% file: figures/toybrains_methods-settings.tex
\begin{tabular}{rrr|lr|rrrrrrr}
\toprule
\multicolumn{3}{c|}{Causal coefficients} & \multicolumn{2}{c|}{DNN Model} & \multicolumn{7}{c}{Ground-truth importance $\bar{\mathrm{Imp}}_{\mathrm{true}}$ ($\pm$ s.d.)} \\
\cmidrule(lr){1-3}\cmidrule(lr){4-5}\cmidrule(lr){6-12}
\shortstack{True Signal\\$\beta_{l_d}$} & \shortstack{Confounder $c_1$\\$\beta_{c_1}$} & \shortstack{Confounder $c_2$\\$\beta_{c_2}$} & \shortstack{BAC (\%)\\{}[min - max]} & Layer & \shortstack{True Signal\\$\mathbf{l}_d$ (\%)} & \shortstack{Confounders\\$c_1$ (\%)} & \shortstack{Confounders\\$c_2$ (\%)} & \shortstack{Encoded\\in Image (\%)} & \shortstack{Encoded\\in Outcome (\%)} & \shortstack{Correlated\\Concept (\%)} & \shortstack{Independent\\Concepts (\%)} \\
\midrule
0.0000 & 0.6190 & 0.8473 & 73 [71--74] & $\mathbf{H}_{L-2}$ & 1 $\pm$ 1 & 11 $\pm$ 4 & 5 $\pm$ 2 & 84 $\pm$ 6 & 0 $\pm$ 0 & 0 $\pm$ 0 & 0 $\pm$ 0 \\
 &  &  &  & $\mathbf{H}_{L-1}$ & 1 $\pm$ 0 & 35 $\pm$ 12 & 18 $\pm$ 11 & 46 $\pm$ 18 & 0 $\pm$ 0 & 0 $\pm$ 0 & 0 $\pm$ 0 \\
 &  &  &  & $\mathbf{H}_{L}$ & 0 $\pm$ 0 & 77 $\pm$ 4 & 21 $\pm$ 4 & 2 $\pm$ 1 & 0 $\pm$ 0 & 0 $\pm$ 0 & 0 $\pm$ 0 \\
\midrule
0.6190 & 0.6190 & 1.7346 & 74 [71--76] & $\mathbf{H}_{L-2}$ & 2 $\pm$ 1 & 6 $\pm$ 2 & 10 $\pm$ 5 & 82 $\pm$ 8 & 0 $\pm$ 0 & 0 $\pm$ 0 & 0 $\pm$ 0 \\
 &  &  &  & $\mathbf{H}_{L-1}$ & 5 $\pm$ 1 & 22 $\pm$ 8 & 35 $\pm$ 7 & 38 $\pm$ 15 & 0 $\pm$ 0 & 0 $\pm$ 0 & 0 $\pm$ 0 \\
 &  &  &  & $\mathbf{H}_{L}$ & 8 $\pm$ 1 & 34 $\pm$ 6 & 56 $\pm$ 5 & 2 $\pm$ 0 & 0 $\pm$ 0 & 0 $\pm$ 0 & 0 $\pm$ 0 \\
\midrule
1.0986 & 0.7082 & 0.0000 & 77 [75--79] & $\mathbf{H}_{L-2}$ & 6 $\pm$ 3 & 13 $\pm$ 6 & 1 $\pm$ 1 & 80 $\pm$ 9 & 0 $\pm$ 0 & 0 $\pm$ 0 & 0 $\pm$ 0 \\
 &  &  &  & $\mathbf{H}_{L-1}$ & 21 $\pm$ 5 & 43 $\pm$ 10 & 1 $\pm$ 1 & 35 $\pm$ 13 & 0 $\pm$ 0 & 0 $\pm$ 0 & 0 $\pm$ 0 \\
 &  &  &  & $\mathbf{H}_{L}$ & 24 $\pm$ 3 & 75 $\pm$ 3 & 0 $\pm$ 0 & 1 $\pm$ 1 & 0 $\pm$ 0 & 0 $\pm$ 0 & 0 $\pm$ 0 \\
\midrule
1.2657 & 0.0000 & 2.1972 & 77 [76--78] & $\mathbf{H}_{L-2}$ & 8 $\pm$ 3 & 1 $\pm$ 0 & 17 $\pm$ 6 & 74 $\pm$ 9 & 0 $\pm$ 0 & 0 $\pm$ 0 & 0 $\pm$ 0 \\
 &  &  &  & $\mathbf{H}_{L-1}$ & 20 $\pm$ 3 & 0 $\pm$ 0 & 48 $\pm$ 8 & 32 $\pm$ 10 & 0 $\pm$ 0 & 0 $\pm$ 0 & 0 $\pm$ 0 \\
 &  &  &  & $\mathbf{H}_{L}$ & 28 $\pm$ 2 & 0 $\pm$ 0 & 71 $\pm$ 1 & 1 $\pm$ 0 & 0 $\pm$ 0 & 0 $\pm$ 0 & 0 $\pm$ 0 \\
\midrule
1.3863 & 0.2007 & 2.1972 & 78 [75--79] & $\mathbf{H}_{L-2}$ & 9 $\pm$ 3 & 1 $\pm$ 0 & 16 $\pm$ 5 & 74 $\pm$ 8 & 0 $\pm$ 0 & 0 $\pm$ 0 & 0 $\pm$ 0 \\
 &  &  &  & $\mathbf{H}_{L-1}$ & 21 $\pm$ 4 & 2 $\pm$ 0 & 44 $\pm$ 10 & 33 $\pm$ 14 & 0 $\pm$ 0 & 0 $\pm$ 0 & 0 $\pm$ 0 \\
 &  &  &  & $\mathbf{H}_{L}$ & 33 $\pm$ 1 & 2 $\pm$ 0 & 64 $\pm$ 1 & 1 $\pm$ 0 & 0 $\pm$ 0 & 0 $\pm$ 0 & 0 $\pm$ 0 \\
\midrule
1.7346 & 0.0000 & 1.2657 & 76 [75--77] & $\mathbf{H}_{L-2}$ & 20 $\pm$ 7 & 0 $\pm$ 0 & 7 $\pm$ 2 & 72 $\pm$ 9 & 0 $\pm$ 0 & 0 $\pm$ 0 & 0 $\pm$ 0 \\
 &  &  &  & $\mathbf{H}_{L-1}$ & 47 $\pm$ 9 & 0 $\pm$ 0 & 16 $\pm$ 3 & 37 $\pm$ 12 & 0 $\pm$ 0 & 0 $\pm$ 0 & 0 $\pm$ 0 \\
 &  &  &  & $\mathbf{H}_{L}$ & 73 $\pm$ 2 & 0 $\pm$ 0 & 26 $\pm$ 1 & 1 $\pm$ 0 & 0 $\pm$ 0 & 0 $\pm$ 0 & 0 $\pm$ 0 \\
\bottomrule
\end{tabular}

%% file: figures/neuro_data-concepts.tex
binge drinking & Outcome & Categorical & Binge (52\%) Non-binge (48\%) \\
\addlinespace[2pt]
age & Demographics & Continuous & 40.0 -- 70.0 ($\mu$=54.3) \\
handedness & Demographics & Categorical & Right-handed (89\%) Left-handed-or-both (11\%) \\
sex & Demographics & Categorical & Male (51\%) Female (49\%) \\
\addlinespace[2pt]
deprivation index & Socioeconomic Status & Continuous & -6.3 -- 9.2 ($\mu$=-1.9) \\
employment status & Socioeconomic Status & Categorical & Employed (72\%) Retired (22\%) Unemployed (6\%) \\
household income & Socioeconomic Status & Categorical & modal category 30\% of n=17396 \\
\addlinespace[2pt]
BMI & Physical Health & Continuous & 13.4 -- 58.7 ($\mu$=26.5) \\
blood pressure systolic & Physical Health & Continuous & 81.0 -- 252.0 ($\mu$=139.6) \\
cholesterol & Physical Health & Continuous & 2.2 -- 12.9 ($\mu$=5.7) \\
has neurological disease & Physical Health & Categorical & False (98\%) True (2\%) \\
physical activities MET & Physical Health & Continuous & 0.0 -- 19278.0 ($\mu$=2416.1) \\
\addlinespace[2pt]
fluid intelligence & Cognitive Tests & Continuous & 0.0 -- 13.0 ($\mu$=6.7) \\
snap reaction time & Cognitive Tests & Continuous & 347.0 -- 1684.0 ($\mu$=589.0) \\
trail making test & Cognitive Tests & Continuous & 0.0 -- 3337.0 ($\mu$=546.4) \\
\addlinespace[2pt]
depression ever & Mental Health & Categorical & True (55\%) False (45\%) \\
ever addicted & Mental Health & Categorical & False (95\%) True (5\%) \\
ever had anxiety & Mental Health & Categorical & False (73\%) True (27\%) \\
frequency of alcohol use & Mental Health & Continuous & 0.0 -- 5.0 ($\mu$=3.2) \\
frequency of cannabis use & Mental Health & Continuous & 0.0 -- 4.0 ($\mu$=0.3) \\
frequency of tobacco use & Mental Health & Continuous & 0.0 -- 4.0 ($\mu$=0.1) \\
happiness & Mental Health & Categorical & modal category 45\% of n=13208 \\
has psychiatric disorder & Mental Health & Categorical & False (94\%) True (6\%) \\
self harm & Mental Health & Categorical & False (84\%) True (16\%) \\
trauma childhood & Mental Health & Categorical & False (91\%) True (9\%) \\
trauma lifethreat & Mental Health & Categorical & False (68\%) True (32\%) \\
\addlinespace[2pt]
date of acquisition & MRI Scanner Settings & Continuous & 2014--2019 \\
imaging site & MRI Scanner Settings & Categorical & Cheadle (71\%) Newcastle (15\%) Reading (15\%) \\
scanner table position & MRI Scanner Settings & Continuous & -1221.0 -- -1020.0 ($\mu$=-1056.7) \\
scanner x position & MRI Scanner Settings & Continuous & -36.3 -- 17.3 ($\mu$=0.5) \\
scanner y position & MRI Scanner Settings & Continuous & 52.0 -- 96.0 ($\mu$=65.4) \\
scanner z position & MRI Scanner Settings & Continuous & -96.8 -- 124.5 ($\mu$=-26.4) \\
\bottomrule